\documentclass{article}
\usepackage{arxiv}

\usepackage{xcolor}
\usepackage[hidelinks]{hyperref}
\usepackage{graphicx}
\usepackage{amsmath}
\usepackage{amssymb}
\usepackage{bm}
\usepackage{array}
\usepackage{caption}
\usepackage{url}
\usepackage{subfigure}
\usepackage[ruled, lined, linesnumbered, commentsnumbered, longend]{algorithm2e}
\SetKwComment{Comment}{$\triangleright$\ }{}
\newcommand{\vect}[1]{\bm{#1}}

\title{Deep Physics Corrector: A physics enhanced deep learning architecture for solving stochastic  differential equations}
\author{ \hspace{1mm}Tushar\\
	Department of Applied Mechanics\\
	Indian Institute of Technology (IIT) Delhi\\
	Hauz Khas - 110 016, New Delhi, India \\
	\texttt{amz218314.iitd@gmail.com} \\
	\And
	\hspace{1mm}Souvik~Chakraborty \\
	Department of Applied Mechanics\\
	Yardi School of Artificial Intelligence (joint appointment)\\
	Indian Institute of Technology (IIT) Delhi\\
	Hauz Khas - 110 016, New Delhi, India \\
	\texttt{souvik@am.iitd.ac.in} \\
}

\begin{document}
\maketitle

\begin{abstract}
We propose a novel gray-box modeling algorithm for physical systems governed by stochastic differential equations (SDE). The proposed approach, referred to as the Deep Physics Corrector (DPC), blends approximate physics represented in terms of SDE with deep neural network (DNN). The primary idea here is to exploit DNN to model the missing physics. We hypothesize that combining incomplete physics with data will make the model interpretable and allow better generalization. The primary bottleneck associated with training surrogate models for stochastic simulators is often associated with selecting the suitable loss function. Among the different loss functions available in the literature, we use the conditional maximum mean discrepancy (CMMD) loss function in DPC because of its proven performance. Overall, physics-data fusion and CMMD allow DPC to learn from sparse data. We illustrate the performance of the proposed DPC on four benchmark examples from the literature. The results obtained are highly accurate, indicating its possible application as a surrogate model for stochastic simulators.
\end{abstract}


\section{Introduction}
Physical systems are governed by the laws of physics and are often represented using either ordinary differential equations (ODE) or partial differential equations (PDE). Theoretically, we can easily obtain a system's response by solving the underlying differential equation(s); however, this is nontrivial in practice as exact analytical solutions of differential equations are often not known. Under such circumstances, one often relies on simulators where the governing physics is solved using numerical techniques such as finite element \cite{reddy2019introduction, chakraborty2014adaptive, ansari2017impact, blatman2010adaptive} and finite volume methods \cite{eymard2000finite}. Unfortunately, such an approach has two major limitations. First, the governing differential equations used are often based on certain assumptions and approximations \cite{brunton2016discovering}. For example, boundary conditions, material properties, and loading conditions are often simplified. Therefore, the results obtained from simulators are often not representative of the actual reality. Secondly, simulators are often computationally expensive and hence, not applicable in real-time. Physics-informed neural networks (PINN) \cite{yang2020physics, mosca2011effectiveness, goswami2020transfer, dwivedi2019distributed}, an alternative to classical solution techniques, also exists in the literature. PINN is computationally efficient and hence, addresses the second issue highlighted above. However, PINN also relies on exact knowledge of physics, which often is not available. Therefore, it is the call of the hour to develop techniques that can address the two limitations highlighted above.

With the rapid development in the field of artificial intelligence and machine learning, data-driven approaches have emerged as a viable alternative in recent times. The idea here is to develop a machine learning model by using data from physics of the system. The most prominent developments in this domain are perhaps the ones based on neural operators \cite{garg2022assessment}. Neural operators enable solving a family of differential equations by learning the mapping between two functional spaces. Popular neural operators existing in the literature include deep operator network \cite{garg2022variational}, Fourier neural operator \cite{rashid2022learning}, graph neural operator \cite{kumar2021grade}, and wavelet neural operator \cite{tripura2022wavelet, thakur2022deep} to name a few. However, neural operators are data hungry and hence, require a large amount of data. Other popular data driven approaches include neural ODE \cite{chen2018neural} and graph attention differential equation \cite{kumar2021grade}, etc. However, the limitations associated with operator learning algorithms hold for these methods as well.

All physical systems have inherent associated randomness \cite{shreve2004stochastic, kloeden1992stochastic} and ignoring these during the design stage can result in an unsafe design. Therefore, it is of utter importance to incorporate uncertainty and investigate its effect on the underlying system. To that end, the governing physics is often represented in terms of a stochastic differential equation (SDE). Use of SDE to model physical phenomenon is common in various domains, including finance, economy, biology, astronomy and engineering to name a few. There exist a rich literature on numerical techniques for solving SDEs; however, the limitations associated with ODEs and PDEs also holds for SDEs as well. Researchers have also investigated the possibilities of exploiting machine learning and deep learning approaches in the past. For example, Chakraborty and Chowdhury developed the stochastic Galerkin approach for solving stochastic heat \cite{chakraborty2016modelling} and stochastic flow \cite{lin2001simple} equations. Thakur and Chakraborty \cite{thakur2022deep} recently developed a deep learning based surrogate model for SDEs. However, data driven models for SDE lack generalization and out-of-distribution predictability, and hence, need repeated retraining.

The limitations associated with both data-driven and physics-based approaches are evident from the discussion above. A possible alternative here is to develop the hybrid model by \textit{data-model fusion}. One of the first work in this work was carried out by Chakraborty \cite{goswami2020transfer} where a transfer learning based multi-fidelity physics informed deep learning framework was developed. Although the framework developed was efficient, it lacked interpretability. Garg \textit{et al.} \cite{garg2022physics} developed an alternative framework by augmenting a machine learning model with known physics. The overall framework involves two stages, with stage 1 being duel Bayesian filtering and stage 2 being Gaussian process. However, this framework is limited to deterministic systems only. Additionally, the two stages are sequential, and hence, the training phase is computationally expensive. Other work on data-model fusion includes \cite{garg2022physics, dwivedi2021distributed}.

The objective of this paper is to develop a novel framework for stochastic dynamical systems governed by SDEs. To that end, we propose Deep Physics Corrector (DPC), a novel approach for solving stochastic dynamic problems. The proposed DPC blends known physics in the form of SDE with deep neural networks (DNN). The primary idea here is to use DNN to model the missing physics. The salient features of the proposed DPC are highlighted below:
\begin{itemize}
    \item \textbf{End-to-end training: }The proposed DPC allows end-to-end training and hence, the training phase is computationally efficient.
    \item \textbf{Physics-data fusion: } Unlike most data-based models, the proposed DPC advocates physics-data fusion. This allows DPC to adopt advantages of both physics-based models and data based models. To be specific, the proposed DPC is partially interpretable, generalizes to new environments, and alleviates the need for exact physics.
    \item \textbf{Surrogate model for stochastic simulators: }DPC can be used as a surrogate model for the stochastic simulator. We note that developing a surrogate model for the stochastic simulator is non-trivial and the proposed DPC provides a viable alternative in this regards.
\end{itemize}
Overall, the proposed DPC is the first of its kind and, to the best of our knowledge, no work on development of such hybrid models for SDEs exists in the literature.

The remainder of the paper is organized as follows. The problem setup is formally described in Section \ref{sec:ps}. Details on the proposed DPC along with the algorithm are provided in Section \ref{sec:dpc}. Numerical experiments carried out are presented in Section \ref{sec:ns}. Finally, Section \ref{sec:concl} presents the concluding remarks.

\section{Problem Statement}\label{sec:ps}
 Consider $\left( {\Omega ,\mathcal{F}, P} \right)$ as the probability space with $\Omega$ being the sample space, $\{{\mathcal{F}_t},0 \le t \le T)\}$ being the natural filtration constructed from sub $\sigma$-algebras of $\mathcal{F}$. With this setup, an $m$-dimensional $n$-factor stochastic differential equation (SDE) driven by $n$-dimensional Brownian motion \{${\bm{B}}_j(t),j = 1, \ldots n$\} is defined as,  
\begin{equation}\label{sdegen}
\begin{array}{l}
d{\bm{X}} = {{\bm f}}\left( {{{\bm{X}}}, \bm \Xi} \right)dt + \sum\limits_{j = 1}^n {{\bm g}_j\left( {{{\bm{X}}}, \bm \Xi} \right)} d{{\bf B}_j}\left( t \right); \quad
{\bm X}(t=t_0)={\bm X}_0; \quad t \in [0,T].
\end{array}
\end{equation}
Here ${\bm{X}} \in {\mathbb{R}^m}$ represents the ${{\mathcal{F}_t}}$-measurable state vector, ${\bm{f}}\left( {{{\bm{X}}}, \bm \Xi} \right):{\mathbb{R}^m} \mapsto {\mathbb{R}^m}$, ${\bm{g}}\left( {{{\bm{X}}},\bm \Xi} \right):{\mathbb{R}^m} \mapsto {{\mathbb{R}}^{m \times n}}$ and ${{\bm{B}}_j}\left( t \right) \in {\mathbb{R}^n}$ are the drift vector, diffusion matrix and Brownian motion respectively. The white noises are generalized derivative Brownian motion i.e. ${\bm{\zeta}} (t) = \dot {\bm{B}}(t)$; this ensures that there exists a corresponding SDE of Eq. \eqref{sdegen}. $\bm \Xi = \{\Xi_1, \Xi_2, \ldots, \Xi_k \}\in\mathbb R^k$ represents the stochastic parameters associated with the system.  
In a compact matrix notation, the SDEs can be expressed as,
\begin{equation}\label{sdeg}
	\begin{array}{l}
		d{{\bm{X}}} = {\bm f}\left( {{{\bm {X}}},\bm \Xi} \right)dt + {\bm g}\left( {{{\bm {X}}}, \bm \Xi} \right)d{\bm B}\left( t \right); \quad {\bm X}(t=t_0)={\bm X}_0; \quad t \in [0,T].
	\end{array}
\end{equation}
In the absence of the stochastic variables $\bm \Xi$ (i.e., when $\bm \Xi$ is deterministic), Eq. \ref{sdeg} can be solved by using stochastic integration schemes such as Euler Maruyama, Milstein and Taylor's 1.5 \cite{kloeden1992stochastic}. On the other hand, Eq. \eqref{sdeg} in the presence of  $\bm \Xi$ can be solved by combining Monte Carlo simulation with stochastic integration. However, in practice, the exact form of the governing SDEs are often not known. To that end, we rewrite Eq. \eqref{sdegen} as follows:
\begin{equation}\label{sdeg2}
	\begin{split}
		d{{\bm{X}}} = &
		\underbrace{{\bm f}_k\left( {{{\bm {X}}},\bm \Xi} \right)dt + {\bm g}_k\left( {{{\bm {X}}}, \bm \Xi} \right)d{\bm B}\left( t \right)}_{\text{known}} +   \underbrace{{\bm f}_{uk}\left( {{{\bm {X}}},\bm \Xi} \right)dt + {\bm g}_{uk}\left( {{{\bm {X}}}, \bm \Xi} \right)d{\bm B}\left( t \right)}_{unknown},
	\end{split}
\end{equation}
where $\bm f_k\left(\cdot\right)$ and $\bm f_{uk}\left(\cdot\right)$ represents the known and unknown portion of the drift vector, such that
\begin{align}
   \bm f\left(\cdot\right) = \bm f_k\left(\cdot\right) + \bm f_{uk}\left(\cdot\right).
\end{align}
Similarly, 
\begin{align}
   \bm g\left(\cdot\right) = \bm g_k\left(\cdot\right) + \bm g_{uk}\left(\cdot\right).
\end{align}
With this setup, the known (approximate) governing SDE can be represented as:
\begin{equation}\label{sdeg3}
		d{{\bm{X}}} = 
		{\bm f}_k\left( {{{\bm {X}}},\bm \Xi} \right)dt + {\bm g}_k\left( {{{\bm {X}}}, \bm \Xi} \right)d{\bm B}\left( t \right); \quad {\bm X}(t=t_0)={\bm X}_0; \quad t \in [0,T]
\end{equation}
However, as the governing SDE in Eq. \eqref{sdeg3} is not exact, the results obtained are also approximate. The order of approximation depends on $f_{uk}\left(\cdot\right)$ and $g_{uk}\left(\cdot\right)$.

Consider that, we have access to high-fidelity data, $\mathcal D = \left\{\bm \xi^{(i)}, \left\{\bm X^{(i)}_{j,1:N_t}\right\}_{j=1}^m \right\}_{i=1}^N$, where $\bm \xi \sim p(\bm \Xi)$ is a realization of $\bm \Xi$. With this setup, it is possible to train a purely data-driven surrogate model \cite{thakur2022deep}
\begin{equation}
\begin{aligned}
\mathcal{W_S} : \mathcal{M}_{\bm\Xi} \times \Omega &\longmapsto \mathbb{R}\\
(\bm{\xi},\omega) &\longmapsto \ \mathcal{W_S(\mathbf{x},\omega)} 
\end{aligned}
\end{equation}
where $ \bm{\xi}\in\mathcal{M}_{\bm \Xi}$ is the input vector; $\mathcal{M}_{\bm \Xi}$ is the input space and 
$\bm{\Omega} \in \mathbb{P}$ is the vector of random variables with $\mathbb{P}$ being the probability space.
However, such models often fail to generalize to unseen environments and hence, has limited applicability. The objective of this paper is to develop a novel algorithm that addresses the challenges associated with the physics-based and data-based approaches discussed above.

\section{Deep physics corrector}\label{sec:dpc}
In this section, we present the detailed formulation of the proposed framework, referred to as the deep physics corrector (DPC). The basic premise here is to augment the known equation of motion with a deep learning model. We hypothesize that combining known physics with a machine learning model will result in models that will generalize better. We start this section by providing the details on the network architecture and loss function and conclude with the overall framework.
\subsection{Architecture design}
The proposed DPC aims to integrate known low-fidelity physics model with sensor data. Accordingly, the architecture consists of two components: (a) the physics module and (b) the deep learning module. In DPC, we augment the two moduli such that the deep learning module is responsible for modeling the missing physics. In this context, we note that the missing physics can either be in the drift part of in the diffusion part. Without any loss of generality, we present the proposed approach by assuming the missing physics to be in the drift. Therefore, we rewrite Eq. \eqref{sdeg2} as follows:
\begin{equation}\label{sdeg2_mod}
	\begin{split}
		d{{\bm{X}}} = &
		\underbrace{{\bm f}_k\left( {{{\bm {X}}},\bm \Xi} \right)dt + {\bm g}_k\left( {{{\bm {X}}}, \bm \Xi} \right)d{\bm B}\left( t \right)}_{\text{known}} +   \underbrace{{\bm f}_{uk}\left( {{{\bm {X}}},\bm \Xi} \right)dt}_{unknown}.
	\end{split}
\end{equation}
Note that in Eq. \eqref{sdeg2_mod}, the unknown portion only consist of the drift component. We rearrange Eq. \eqref{sdeg2_mod} and place a neural network prior on the unknown physics. Accordingly,
\begin{equation}\label{sdeg2_modnn}
		d{{\bm{X}}} = 
		\underbrace{{\bm f}_k\left( {{{\bm {X}}},\bm \Xi} \right)  + \mathcal N_N (\bm X, \bm \Xi, \bm Z; \bm \theta )}_{\text{drift}} +   \underbrace{{\bm g}_k\left( {{{\bm {X}}}, \bm \Xi} \right)d{\bm B}\left( t \right)}_{\text{diffusion}},
\end{equation}
where $\mathcal N_N (\bm X, \bm \Xi, \bm Z; \bm \theta )$ represents neural network parameter, parameterized by $\bm \theta$. Note that a generative neural network is used in Eq. \eqref{sdeg2_modnn} and hence, additional variables $\bm Z$ are introduced. Generative neural network allows the model to track the uncertainty due to limited data. Additionally, it also helps in avoiding overfitting. Following the usual norm, we model $\bm Z \sim \mathcal N(\bm 0, \mathbf I)$ as standard normal variable \cite{thakur2022deep}.

\begin{figure}[ht!]
    \centering
    \subfigure[DPC cell]{
    \includegraphics[width=0.75\textwidth]{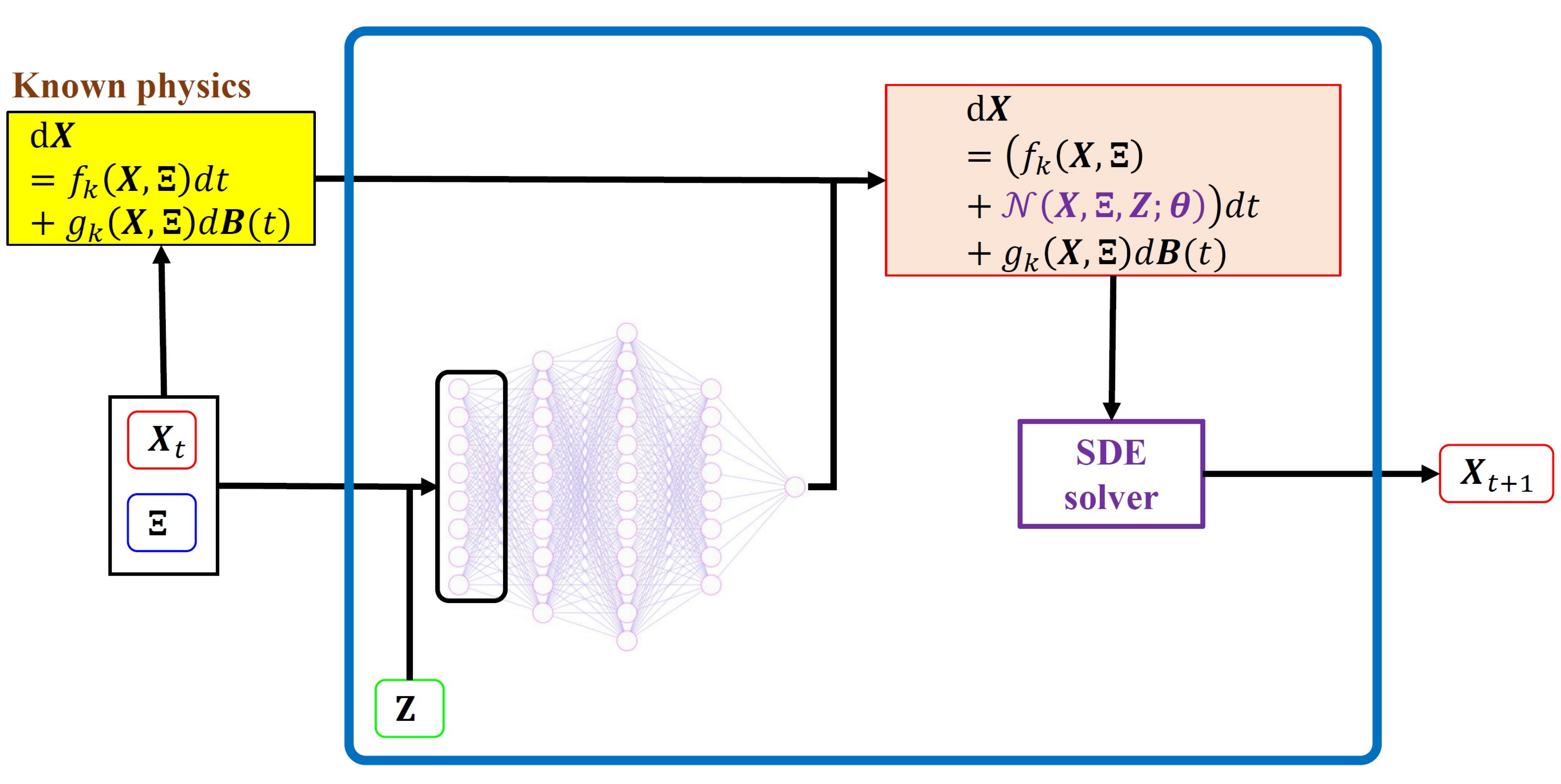}}
    \subfigure[Overall architecture]{
    \includegraphics[width=\textwidth]{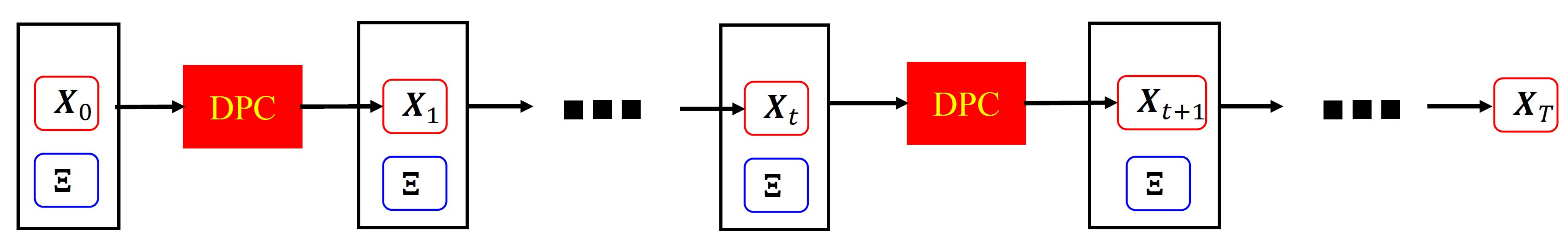}}
    \caption{Network architecture of the proposed DPC. (a) one unit cell of DPC (the blue box) and (b) the overall architecture developed using DPC cell. The DPC cell in the overall architecture is shown in red.}
    \label{fig:dpc}
\end{figure}

For designing the neural network architecture, we use stochastic integration scheme to express Eq. \eqref{sdeg2_modnn} in discretized form. This yields
\begin{equation}\label{eq:em_pa}
 {\bm X_{t+1} = \bm X_{t} + \left\{ {\bm f}_k\left( {{{\bm {X}}_t},\bm \Xi} \right)  + \mathcal N_N (\bm X_t, \bm \Xi, \bm Z; \bm \theta ) \right\}\cdot\Delta t  +  \left\{{\bm g}_k\left( {{{\bm {X}}_t}, \bm \Xi} \right)\right\}\cdot \Delta \bm B_t}   ,
\end{equation}
where $\Delta t$ represents the time-step and $\Delta \bm B_t \sim \mathcal N \left( \bm 0, \sqrt{\Delta t}\cdot \mathbf I \right)$ is a normal distributed variable. $\bm X_{t+1}$ is the response at time-step $t+1$. Using Eq. \eqref{eq:em_pa}, we can start with a given initial condition and time-march to obtain the response at any time. The neural network architecture used in this study is designed by following Eq. \eqref{eq:em_pa}. A schematic representation of the neural network architecture is shown in Fig. \ref{fig:dpc}.

\noindent \textbf{Remark 1: }Although Euler Maruyama scheme has been used to develop the proposed DPC algorithm, it can be trivially extended to incorporate higher order stochastic integration scheme (e.g., Taylor's 1.5 scheme for example). However, this might result in added computational cost.

\subsection{Loss function}
For training the proposed DPC, selecting a suitable loss function is essential. Two of the most popular loss functions existing in the literature are perhaps the mean squared and cross entropy loss functions. While the former is often used in regression, the latter is used while solving classification problems. Unfortunately, neither of these can be used in conjunction with the proposed DPC because the DPC is developed for systems governed by SDE; hence, the output is probabilistic and represented using the probability density function. Therefore, a suitable loss function is one that attempts to minimize distance between two distributions. A rich literature on distance metric also exists in the literature with Kullback-Liebler (KL) divergence {\cite{kullback1997information}} and maximum mean discrepancy (MMD) \cite{gretton2012kernel} being two of the popular choice. This study, uses a modified version of MMD loss function, referred to as the conditional maximum mean discrepancy (CMMD) {\cite{li2015generative}}. Details on CMMD and its utilization in training DPC is provided next.
\subsubsection{RKHS and MMD}
The concept of CMMD is rooted in Reproducing Kernel Hilbert Space (RKHS) and kernel embedding. Therefore, we start our discussion with the same. A Hilbert space is defined as the space of inner products, which is complete and separable with respect to the norm defined by the inner products. A kernel, $k(\cdot,\cdot)$, is said to be a reproducing kernel in a Hilbert space $\mathcal{H}$, if $\forall{f}\in \mathcal{H}, f(x) = \langle k(x,\cdot),f(\cdot)\rangle$. An RKHS is a Hilbert space $H$ with a reproducing kernel spanned within $H$, alternatively, it is the space of functions with all evaluation functions bounded and linear \cite{Peter:2008}. MMD is a kernel- based method to compare distance between two distributions. Assuming  $\vect{Y}^t = \bigl\{t^t_{i}\bigr\}_{i=1}^N \sim P_{\vect{Y}^t}$ and $\vect{Y}^p = \bigl\{y^p_{j}\bigr\}_{j=1}^M \sim P_{\vect{Y}^p}$, MMD is represented as
\begin{equation}
\operatorname{MMD}[\mathcal{F},P_{\vect{Y}^t},P_{\vect{Y}^p}] = \sup_{f\, \in\,\mathcal{F}}\left(\mathbb{E}_{\vect{Y}^t}\left[f(\vect{Y}^t)\right] - \mathbb{E}_{\vect{Y}^p}\left[f(\vect{Y}^p)\right]  \right),
\label{equa3}
\end{equation}
where $\mathcal{F}$ is a class of functions. \cite{gretton2012kernel} showed that Eq.  \eqref{equa3} can be solved in the closed form if $\mathcal{F}$ is specified to be an RKHS $\mathcal{K}$ and hence,
\begin{equation}
\operatorname{MMD}(\mathcal{F}, p, q)=\left\|\mu_{p}-\mu_{q}\right\|_{\mathcal{K}}^{2},
\label{equa4}
\end{equation}
where $\mu_{p},\mu_{q} \in \mathcal{K}$. However, instead of Eq. \eqref{equa4}, we use squared difference between the kernel mean embeddings,
\begin{equation}
\widehat{\mathcal{L}}_{\mathrm{MMD}}^{2}=\left\|\frac{1}{N} \sum_{i=1}^{N} \phi\left(y^t_{i}\right)-\frac{1}{M} \sum_{j=1}^{M} \phi\left(y^p_{i}\right)\right\|_{\mathcal{K}}^{2}.
\label{equa5}
\end{equation}
With suitable choice of $\phi$ in Eq. \eqref{equa5}, it is possible to match higher order moments \cite{gretton2012kernel}. Using kernel trick, Eq.  \eqref{equa5} can further be simplified as
\begin{equation}
\widehat{\mathcal{L}}_{\mathrm{MMD}}^{2}=E\left[k\left(\vect{X}, \vect{X}^{\prime}\right)-2 k(\vect{X}, \vect{Y})+k\left(\vect{Y}, \vect{Y}^{\prime}\right)\right]
\label{equa6}
\end{equation}
\subsubsection{CMMD}
CMMD is a natural extension of MMD and is used when dealing with conditional distributions. Training DPC involves matching the conditional distribution of the predicted response with the target response and hence, CMMD becomes a natural choice.

Consider $\mathcal D^t = \left(\bm \xi^{(i)}, \bm X_t^{(i)} \right)_{i=1}^M$ and $\mathcal D^p = \left(\bm \xi^{(j)}, \bm X_p^{(j)} \right)_{j=1}^N$ to be target and predicted datasets. Since we are dealing with SDE, the output samples $\bm X_t \sim p(\bm X | \bm \xi^*)$ and $\bm X_p \sim q(\bm X | \bm \xi^*)$ can be thought of as samples drawn from the target conditional distribution  $p(\bm X | \bm \xi^*)$ and predicted conditional distribution $q(\bm X | \bm \xi^*)$ respectively. Accordingly, the CMMD loss function is represented as
\begin{equation}
    \mathcal L_{\text{CMMD}}^2 = \left\| C_{Y^t|\bm \xi} - C_{Y^p|\bm \xi}  \right\|^2_{\mathcal K \otimes \mathcal K},
\end{equation}
where $C_{Y^t|\bm \xi}$ is the conditional embedding operator. In practice, we utilize the empirical estimate of conditional embedding operators, with $\widetilde{\mathbf{K}} = \mathbf{K} + \lambda\mathbf{I} $, where $\mathbf{K}$ is the gram matrix, used to compute the discrepancy between the two above given distributions, using the kernel trick. The empirical estimate is as follows:
\begin{equation}\label{eq:cmmd_fin}
\begin{aligned}
\widehat{\mathcal{L}}_{\mathrm{CMMD}}^{2} &=\left\|\Phi_{d}\left(\mathbf{K}_{d}+\lambda \mathbf{I}\right)^{-1} \Upsilon_{d}^{\top}-\Phi_{s}\left(\mathbf{K}_{s}+\lambda \mathbf{I}\right)^{-1} \Upsilon_{s}^{\top}\right\|_{\mathcal{K} \otimes \mathcal{H}}^{2} \\
&=\operatorname{Tr}\left(\mathbf{K}_{d} \widetilde{\mathbf{K}}_{d}^{-1} \mathbf{L}_{d} \widetilde{\mathbf{K}}_{d}^{\prime}\right)+\operatorname{Tr}\left(\mathbf{K}_{s} \widetilde{\mathbf{K}}_{s}^{-1} \mathbf{L}_{s} \widetilde{\mathbf{K}}_{s}^{-1}\right)-2 \cdot \operatorname{Tr}\left(\mathbf{K}_{s d} \widetilde{\mathbf{K}}_{d}^{-1} \mathbf{L}_{d s} \widetilde{\mathbf{K}}_{s}^{-1}\right),
\end{aligned}
\end{equation}
where $\gamma_d$ and $\gamma_s$ represents feature matrix for input variables, for data-sets of subscripts $d$ and $s$ respectively and $\phi_d$ and $\phi_s$ represents feature matrix for output variables, for data-sets of subscripts $d$ and $s$ respectively. Further, $\mathbf{K}_s = {\gamma_s}^T \gamma_s$, $\mathbf{K}_d = {\gamma_d}^T \gamma_d$ are defined as the gram matrices for the input variables and  $\mathbf{L}_s = {\phi_s}^T \phi_S$, $\mathbf{L}_d = {\phi_d}^T \phi_d$ are defined as the gram matrices for the output variables, with $\mathbf{K}_{sd} = {\gamma_s}^T \gamma_d$, $\mathbf{L}_{ds} = {\phi_d}^T \phi_d$ are defined to be the gram matrices containing the dot product of feature matrices of the two distributions \cite{thakur2022deep}. We utilize Eq. \eqref{eq:cmmd_fin} as loss function in the proposed DPC.

\subsection{Algorithm}
Having discussed the network architecture and the loss function, we proceed with discussing the proposed algorithm. The code accompanying the proposed approach is developed using \texttt{PyTorch} and Adam optimizer is used to train the model. Details on the algorithm are discussed in Algorithm \ref{alg:framework}.
\begin{algorithm}[ht!]
\caption{Algorithm for the proposed DPC}\label{alg:framework}
\SetKwInOut{KwIn}{Input} 
\SetKwInOut{KwOut}{Output}
\KwIn{Training data: $\mathcal D = \left\{\bm \xi^{(i)}, \left\{\bm X^{(i)}_{j,1:N_t}\right\}_{j=1}^m \right\}_{i=1}^N$, initial condition: $\bm X_0$, learning rate: $\eta$, time-step: $\Delta t$, and learning-rate scheduler}
Initialize network parameters and hyper-parameters $\bm \theta, \lambda, \beta$.\\
\For{e = 1 to number of epochs}{
     \For{b = 1 to $N_b$}{
          {Draw a mini-batch $\mathcal{A} =  \left\{\bm \xi^{(i)},\left\{ \bm X_{j,1:N_t}^{(i)}\right\}_{j=1}^m \right\}_{i=1}^n$ from $\mathcal D$}\\
          \For{t = $0$ to $N_t$}{
              {$\bm Z \sim \mathcal N(\bm 0, \mathbf I)$}\\
              {$\left[f_{pseudo},g_{pseudo}\right] \longleftarrow \mathcal{N_N} (\bm X, \bm \Xi, \bm Z; \bm \theta)$}\\
            {$\bm X_p(t+1) = \bm X_p(t) + \left\{ {\bm f}_k\left( {{{\bm {X}}_p},\bm \Xi} \right)  + f_{pseudo}\right\}\cdot\Delta t  +  \left\{{\bm g}_k\left( {{{\bm {X}}_p}, \bm \Xi} \right) + g_{pseudo}\right\}\cdot \Delta \bm B_t$ }\\
            {Append $\bm{X_p}$ with $\bm{X_p(t+1)}$ marching in time}}
      {Compute the training loss, $\widehat{\mathcal{L}}_{\mathrm{CMMD}}^{2}$ by using $\bm \xi^{(i)}$, $\left\{ \bm X_{j,1:N_t}^{(i)} \right\}_{j=1}^m$ and  $\left\{ \bm X_{p,j,1:N_t}^{(i)} \right\}_{j=1}^m$\Comment*[r]{Eq. \eqref{eq:cmmd_fin}}}
      {Calculate gradient $\frac{\partial{\widehat{\mathcal{L}}}_{\mathrm{CMMD}}^{2}}{\partial \bm \theta_p}$ and update $\bm \theta_p = \left[\bm \theta, \lambda, \beta \right]$ using gradient based Adam optimizer}
    }}
\KwOut{Learnt the parameter $\bm \theta$, $\lambda$ and $\beta$.}
\end{algorithm}

Once the network is trained, we can predict the response corresponding to a given input $\bm \xi^*$ and initial condition $\bm X_0^*$ by using the training DPC network. In this paper, we use the proposed DPC as a surrogate model for stochastic simulators.

\section{Numerical Examples}\label{sec:ns}
In this section, we present four examples to illustrate the efficacy of the proposed DPC. We select the example problems from different fields i.e., finance(stock prices), physical sciences, medical (epidemiology), and advanced dynamics, to illustrate the vast application domain of the proposed approach. For all the problems, the objective is to predict the probability density function (PDF) of the response variables at a given time $t$. Results obtained using the proposed DPC are benchmarked against results obtained using Monte Carlo simulation (MCS) results (ground truth),  purely data-driven deep learning framework and  purely physics-driven approach with the known physics (Euler Maruyama). Case studies involving (a) missing physics in drift and (b) missing physics in diffusion have been considered. Additional case studies involving the performance of the proposed approach with an increase in the number of training samples and missing physics in both drift and diffusion have also been carried out.

For training the proposed DPC, we selected a neural network architecture involving 5 hidden layers with 64, 256, 512, 256 and 64 nodes. The Exponential linear unit (ELU) activation function was used for all but the last layer. The final layer has no activation function. The dimensionality of the intrinsic variable is ten, $\bm Z \in \mathbb R^{10}$, and it is sampled from standard Gaussian distribution. A Squared exponential kernel function has been used within the CMMD loss function. For all the problems, we have tested the performance of the model for two cases:
\begin{itemize}
    \item \textbf{Case 1:} We test the predictive capability on different values of $\bm \xi^*$ but the prediction time $t^*$ is within the training window. This is the setup used in most of the existing reliability engineering and uncertainty quantification literature.
    \item \textbf{Case 2:} Here, the prediction $t^*$ outside the training window.
\end{itemize}
Other problem-specific details are provided with each example separately.

\subsection{Example 1: Black-Scholes}\label{subsec:eg1}
In this example, we consider the Black-Scholes equation {\cite{black2019pricing}}. This is a popular SDE and has applications in stock prices and future contracts. The equation for the same looks like:
\begin{equation}
d S_{t}=x_{1} S_{t} d t+x_{2} S_{t} d B_{t},
\end{equation}
where $S_t$ represents the output response at any given time, $x_1$ represents the expected output in time $dt$,  $x_2$ gives the measure of volatility around that expected output $x_1$, and $B_t$ is the white noise that imparts the stochastic nature to the process.
We consider $S_t$, at any time $t$, as the quantity of interest (QoI). The values of input parameters $x_1$ and $x_2$ are randomly sampled from the uniform distributions $x_1 \sim \mathcal{U}(0,0.1)$ and $x_2 \sim \mathcal{U}(0,0.4)$. The initial value of stock price $S_0$ is fixed at $1$.

First, we consider the case where the known physics is of the following form:
\begin{equation}\label{eq:bs_drift}
    dS = dt + x_2 S dB_t.
\end{equation}
This corresponds to incomplete/missing physics in the drift component of the SDE. To accommodate for the incomplete physics, we also have access to high-fidelity sensor data, $\mathcal D = \left\{\bm \xi^{(i)}, \left\{S^{(i)}_{j,1:N_t}\right\}_{j=1}^m \right\}_{i=1}^N$ with $m$ being the number of replication per training sample, $N$ being the number of training samples, and $N_t$ being the time-step until which data is available. Here, $\bm \xi^{(i)} = \left[x_1^{(i)}, x_2^{(i)} \right]$, $m = 50$, $N = 40$, and $N_t = 100$. We have considered a time-step $\Delta t =  0.001$ sec and hence, we have data until $t = 0.1$ sec. The model is trained with a constant learning rate of 1e-5, using gradient-based Adam optimizer. The objective here is to use the proposed DPC to solve the Black-Scholes SDE by using Eq. \eqref{eq:bs_drift} and the high-fidelity data $\mathcal D$.

\begin{figure}[t]
    \centering
    \subfigure[$t=0.022$ sec]{ 
      \includegraphics[width=0.48\textwidth]{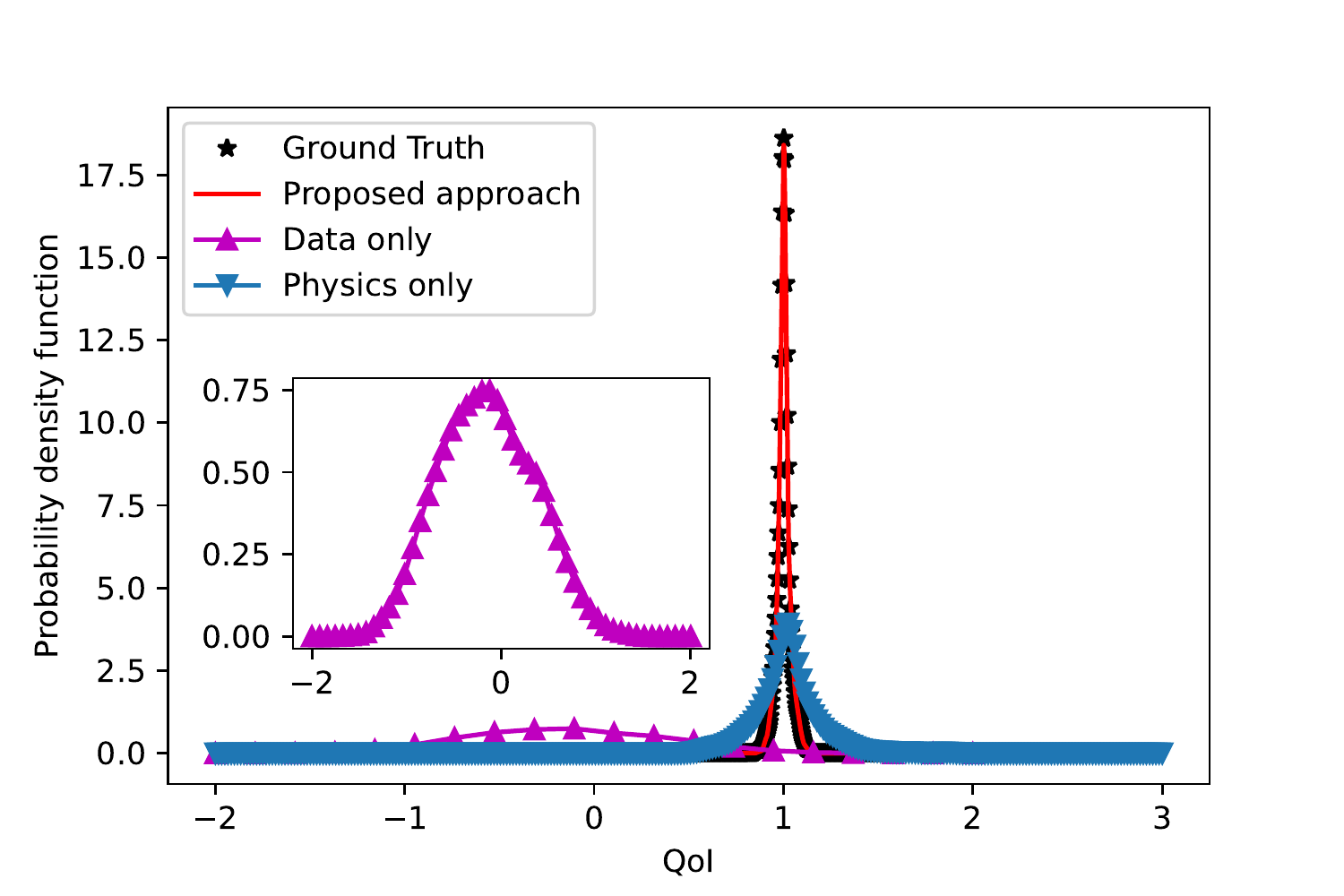}
      \label{fig:BS_drift_a}}
    \subfigure[$t=0.67 sec$]{
      \includegraphics[width=0.48\textwidth]{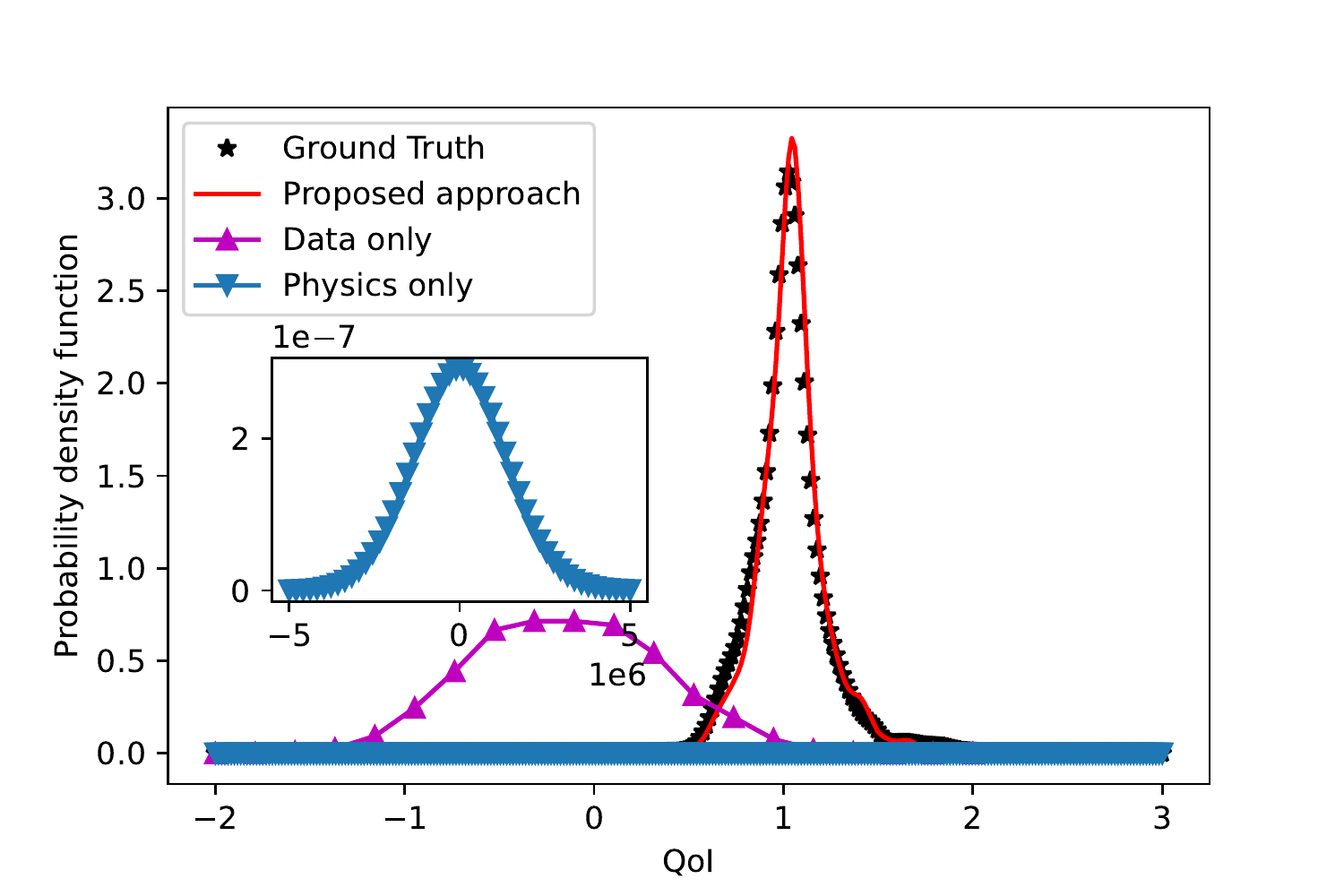}
      \label{fig:BS_drift_b}}
    \caption{PDF comparison of output data-set at a specific time, computed from different models with that of the ground truth, for the case where the drift term is imprecisely known a) PDF comparison within the trained regime b) PDF comparison for extrapolated data-set (prediction) outside the training regime. The model is trained for the first $0.1$ sec }
    \label{fig:BS_drift}
\end{figure}

Fig. \ref{fig:BS_drift_a} shows the PDF of the QoI obtained using different methods at $t = 0.022$ sec. We observe that the PDF obtained using the proposed DPC matches  excellently with that of the ground truth. On the other hand, the \textit{data-only} approach,  fails to capture the temporal evolution of the QoI as indicated by the mismatch in the PDF. The performance of the \textit{physics-only} model is better and it correctly captures the mean response. However, it overestimates the  standard deviation, as indicated by the flat PDF. This can be attributed to the fact that the drift is considered to be constant in Eq. \eqref{eq:bs_drift} and hence, the diffusion term starts dominating and results in a higher standard deviation. 

Fig. \ref{fig:BS_drift_b} shows the response PDFs at $t=0.67$ sec. Since the model was trained only up to t=$0.1$ sec, this prediction is way ahead of the  training window. We observe that the proposed DPC yields excellent results, with the PDF of QoI matching almost exactly with the ground truth. This clearly indicates the generalization capability of the proposed approach. On the other hand, the \textit{physics-only} and \textit{data-only} models, fail to capture the PDF of the QoI. The failure of the data-driven approach is attributed to the lack of generalization of the \textit{data-only} model. The \textit{physics-only} model fails because the effect of mis-specified physics becomes prominent with an increase in time. 

Next, we consider the case with mis-specified physics in the diffusion. The known physics in this case is represented as
\begin{equation}
    dS = x_1 Sdt + dB_t.
\end{equation}
Other setups including the training data size are kept unchanged. 
\begin{figure}[t]
    \centering
    \subfigure[$t=0.098$ sec]{ 
      \includegraphics[width=0.48\textwidth]{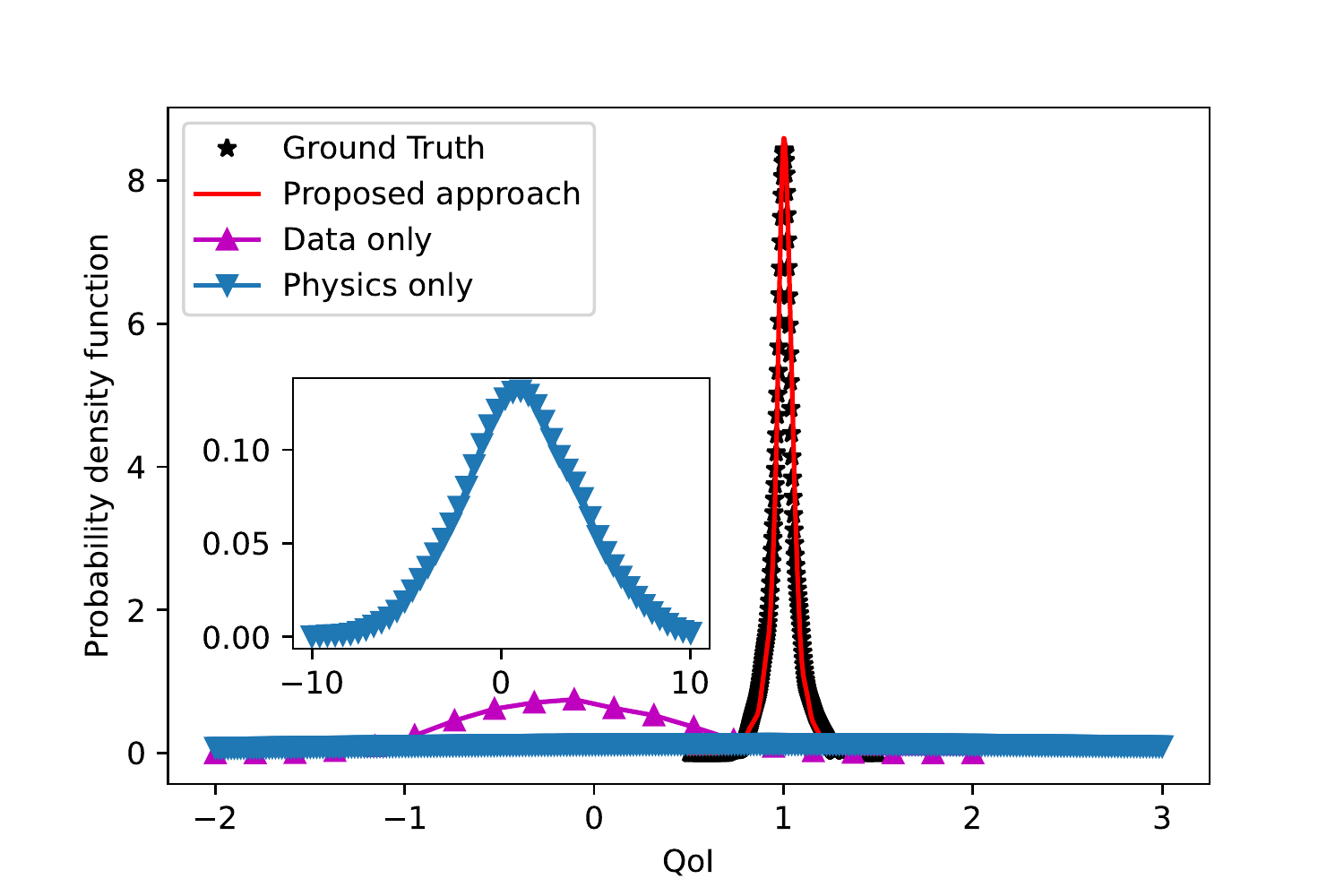}
      \label{fig:BS_diff_a}}
    \subfigure[$t=0.7$ sec]{
      \includegraphics[width=0.48\textwidth]{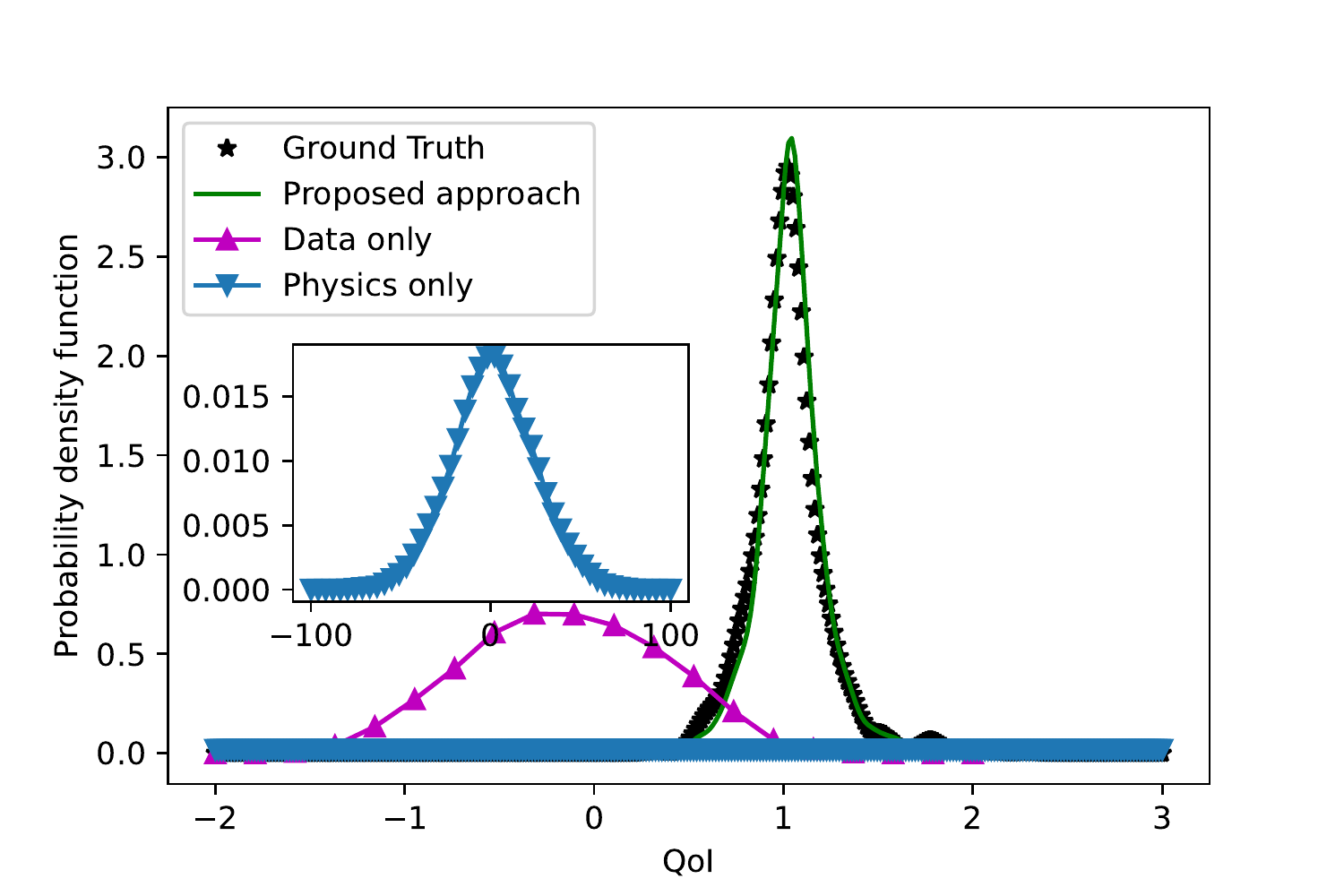}
      \label{fig:BS_diff_b}}
    \caption{PDF comparison of output data-sets at a specific time, computed from different models with that of the ground truth, for the case where the diffusion term is imprecisely known a) PDF comparison within the trained regime b) PDF comparison for predicted data-sets (outside the training window). The model is trained for the first $0.1$ sec }
    \label{fig:BS_diff}
\end{figure}

\begin{figure}[t]
    \centering
    \subfigure[$t=0.068$ sec]{ 
      \includegraphics[width=0.48\textwidth]{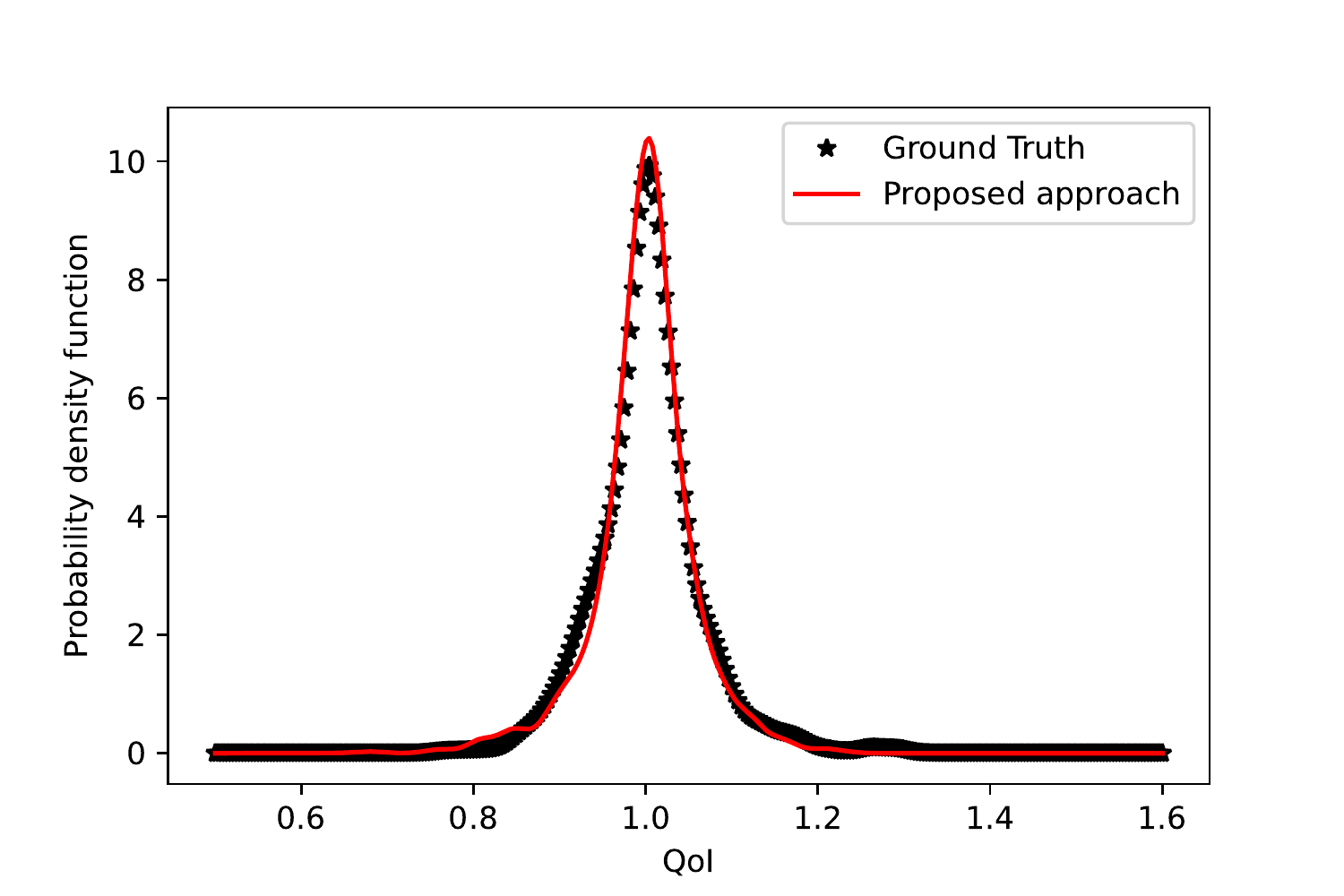}
      \label{fig:BS_both_a}}
    \subfigure[$t=0.26$]{
      \includegraphics[width=0.48\textwidth]{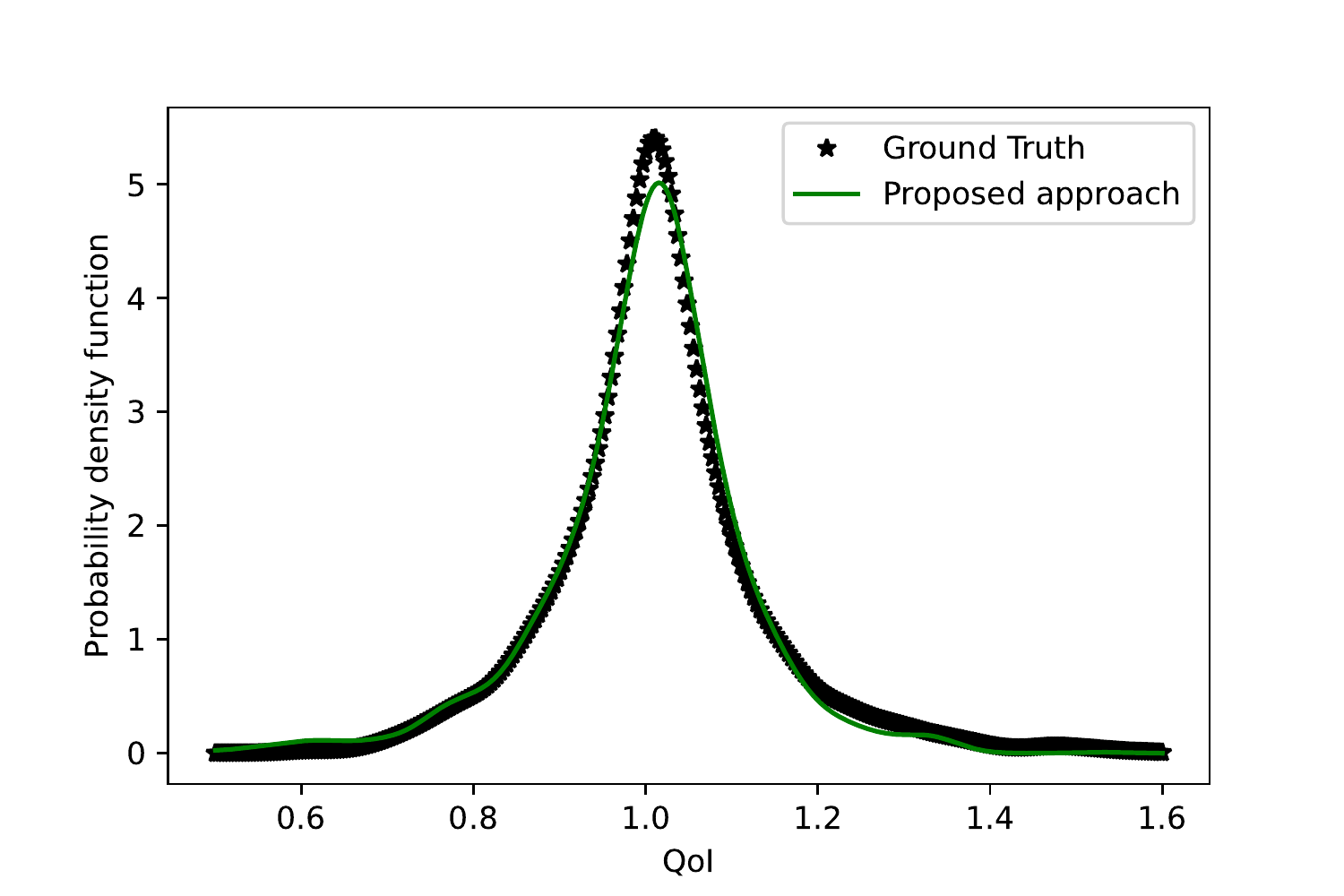}
      \label{fig:BS_both_b}}
    \caption{PDF comparison of output data-sets at a specific time, computed from different models with that of the ground truth, for the case where both the drift and the diffusion terms are imprecisely known a) PDF comparison within the trained regime b) PDF comparison for predicted data-sets (outside the training window. The model is trained for first $0.1$ sec }
    \label{fig:BS_both}
\end{figure}

Figs. \ref{fig:BS_diff_a} and \ref{fig:BS_diff_b} show the PDFs of QoI at $t = 0.098$ sec (within the training window) and $t = 0.7$ sec (outside the training window). For both cases, the proposed DPC yields highly accurate results with the response PDF matching exactly, with the ground truth. The \textit{physics-only} model failed to capture the output distribution, with a slight offset mean, and a completely flat curve. The shift in mean can be attributed to the accumulated error during time marching, which is present due to incorrect diffusion term. The \textit{data-only} model also fails to capture the PDF of QoI for both cases. This indicates poor generalization of solely data-driven approaches. Lastly, we consider the case with mis-specified drift and diffusion. Figs. \ref{fig:BS_both_a} and \ref{fig:BS_both_b} show the PDF of QoI obtained using the proposed DPC. We observe that the results obtained using the proposed approach have an excellent match with the ground truth. This indicates that the proposed approach can also be used when both drift and diffusion components are mis-specified.

One of the important factors governing the performance of the proposed approach is the number of training samples. Therefore, we conclude this section by presenting a case study illustrating the performance of the proposed DPC with an increase in training samples. To that end, we use 
Hellinger distance to quantify the predictive error. Considering $P(y(t))$ and $Q(y(t))$ to be two PDFs, Hellinger distance can be mathematically represented as:
\begin{equation}\label{eq:hellinger}
  H(P(y(t)),Q(y(t))) =  \frac{1}{\sqrt{2}} \left\|\sqrt{P(y(t))}-\sqrt{Q(y(t))}\right\|_2 
\end{equation}
We use normalized time-averaged Hellinger distance in this work,
\begin{align}\label{eq:eps_n}
    \epsilon_n = \frac{\epsilon}{\epsilon_0},
\end{align}
where 
\begin{align}
    \epsilon = \mathbb E \left[H(P(y(t)),Q(y(t)))\right].
\end{align}
$\mathbb E \left[ \cdot \right]$ here is the expectation operator and $\epsilon_0$ in Eq. \eqref{eq:eps_n} is the normalizing factor. Here we consider $\epsilon$ with 40 training samples as $\epsilon_0$. This allows us to track the change in error with the increase in training samples.

\begin{figure}[t]
    \centering
    \includegraphics[width=0.55\textwidth]{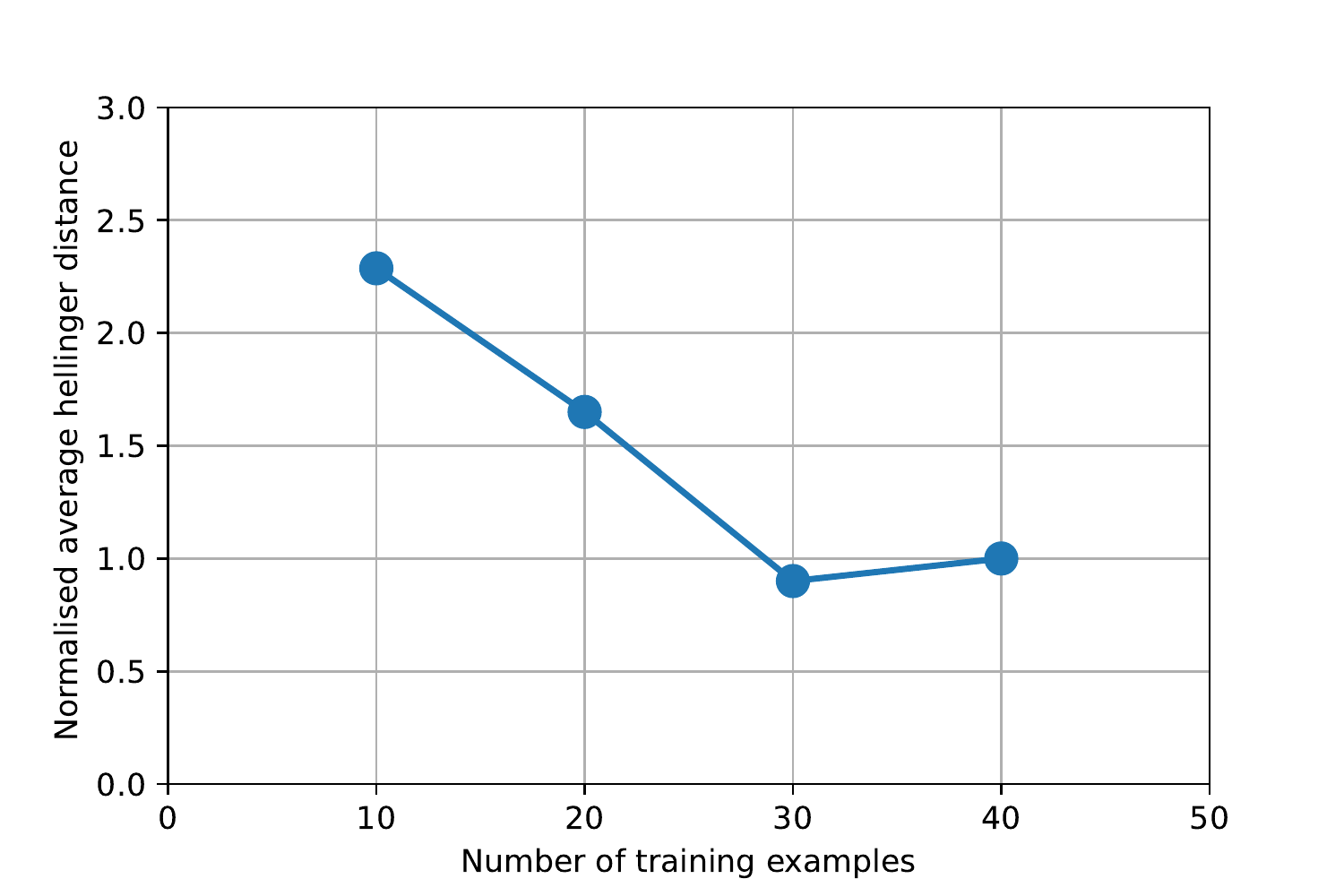}
    \caption{Time averaged Hellinger distance between computed PDF and ground truth PDF with decreasing number of training samples. The computation is done for the case with incomplete drift term in the SDE}
    \label{fig:hellinger}
\end{figure}

Fig. \ref{fig:hellinger} shows the variation of $\epsilon_n$ with the increase in training samples. We observe that the error metrics are similar at $N=30$ and $N=40$, indicating that convergence has been achieved.

\subsection{Example 2: Modified Ornstein–Uhlenbeck process} \label{subsec:eg2}
As the second example, we consider the modified Ornstein–Uhlenbeck process suggested in \cite{zhu2020replication}. The equation for the same is as follows:
\begin{equation}
    d Y = (x_1 - Y)dt + (\nu Y + 1)x_2 dB_t
\end{equation}
where $Y(t)$ is the response at time $t$, $x_1 \sim \mathcal{U}(0.9,2)$ and $x_2 \sim \mathcal{U}(0.1,1)$ are the stochastic parameters and $B_t$ represents the white noise as before. We have taken $\nu = 0.2$, for this problem The initial value is fixed at $1$ and we consider $Y(t)$ as the QoI.

First, we consider the case where the known physics is of the following form:
\begin{equation}\label{eq:ou_drift}
   d Y = dt + x_2(\nu Y + 1) dB_t
\end{equation}
This corresponds to incomplete/missing physics in the drift component of the SDE. To accommodate for the missing physics, we also have access to high-fidelity sensor data, $\mathcal D = \left\{\bm \xi^{(i)}, \left\{Y^{(i)}_{j,1:N_t}\right\}_{j=1}^m \right\}_{i=1}^N$ as before. Here, $\bm \xi^{(i)} = \left[x_1^{(i)}, x_2^{(i)} \right]$, $m = 50$, $N = 40$, and $N_t = 100$. We have considered a time-step $\Delta t =  0.001$ sec and hence, we have data until $t = 0.1$ sec. The model is trained with a  learning rate of $1\times 10^{-4}$, which decreases by a factor of 0.95 after every 50 epochs. For training the model, ADAM optimizer. The objective here is to use the proposed DPC to solve the governing SDE by using Eq. \eqref{eq:ou_drift} and the high-fidelity data $\mathcal D$.



\begin{figure}
    \centering
    \subfigure[$t=0.03$ sec]{ 
      \includegraphics[width=0.48\textwidth]{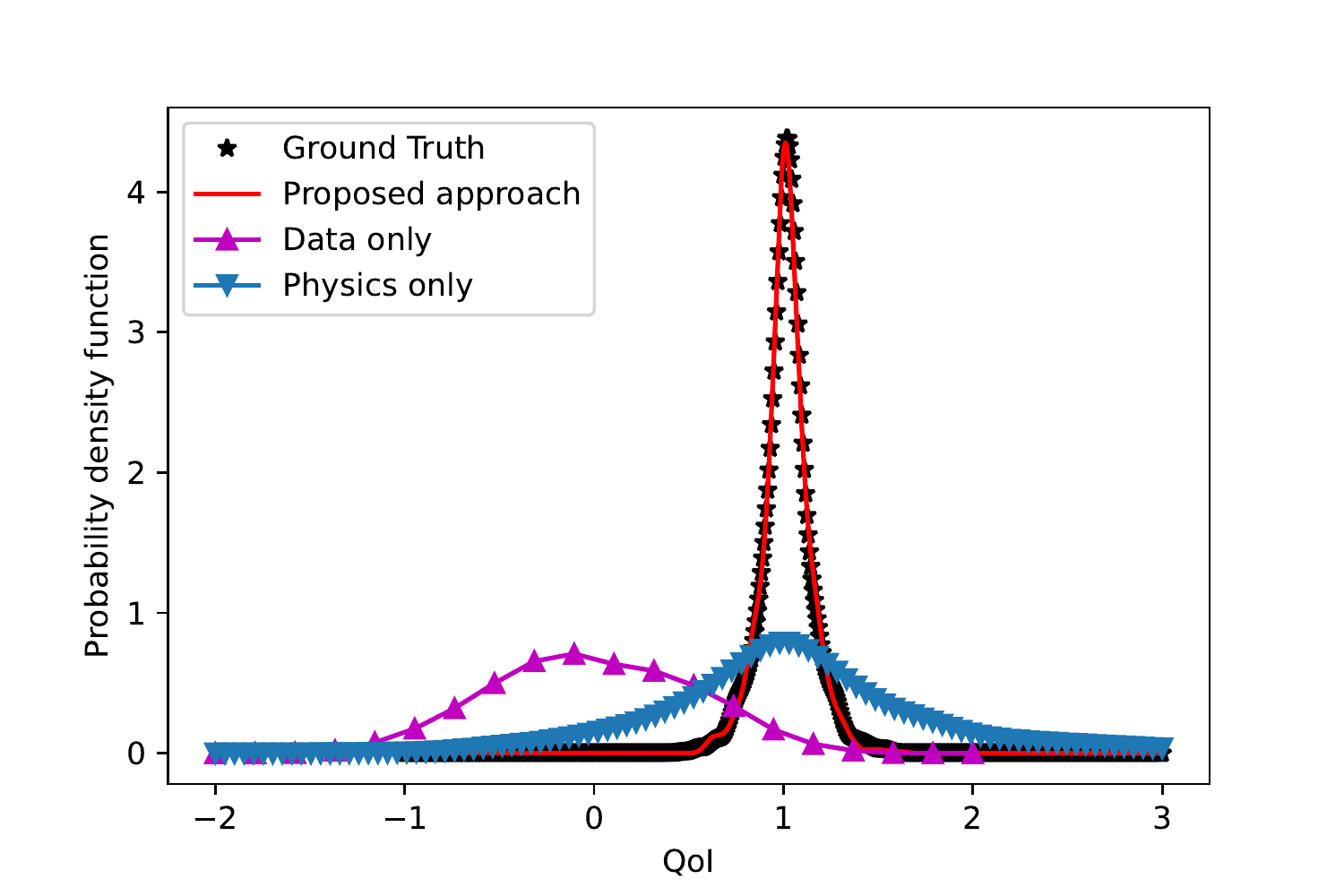}
      \label{fig:SDE2_drift_a}}
    \subfigure[$t=0.2$ sec]{
      \includegraphics[width=0.48\textwidth]{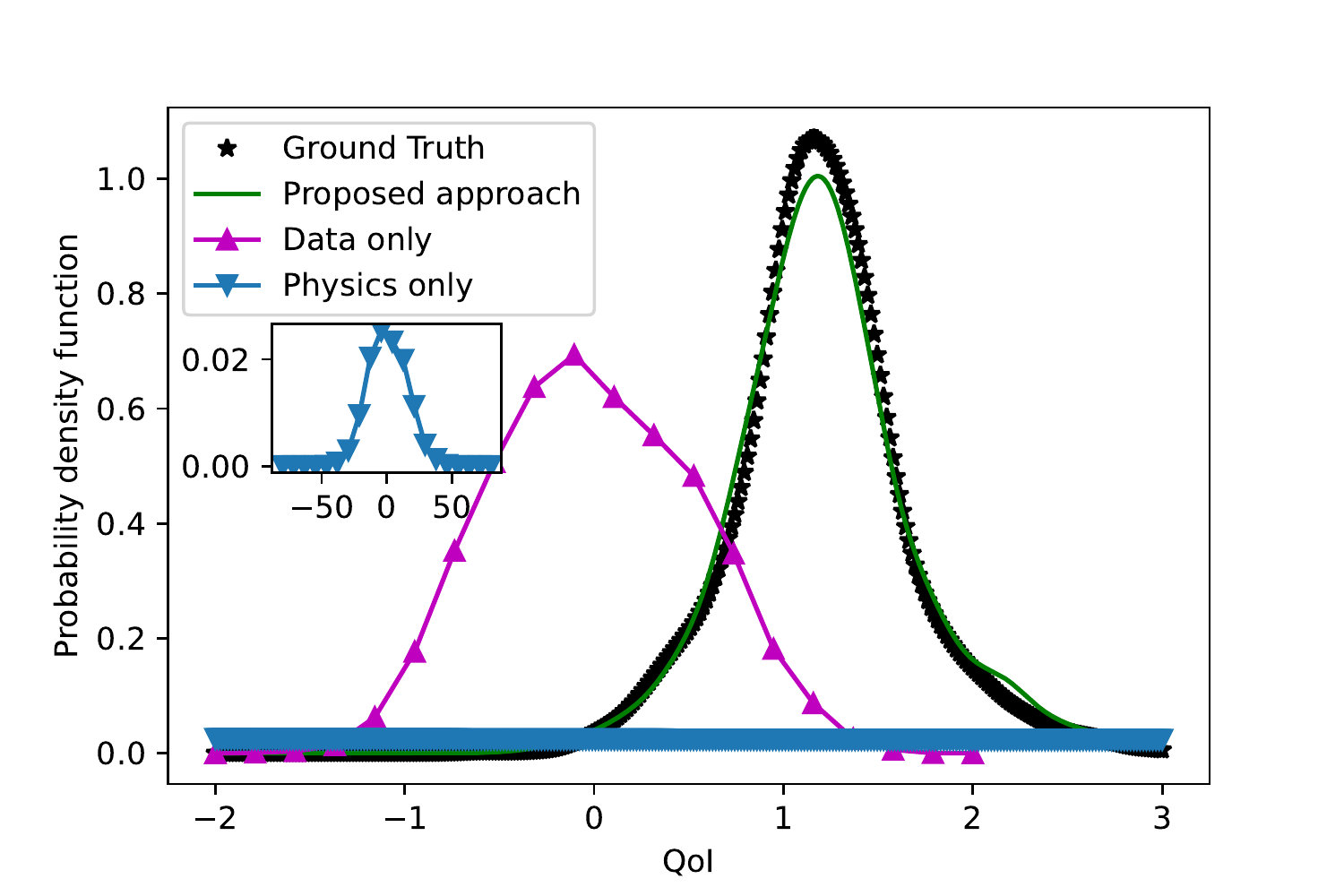}
      \label{fig:SDE2_drift_b}}
    \caption{PDF comparison of output data-sets at a specific time, computed from different models with that of the ground truth, for the case where the drift term is imprecisely known a) PDF comparison within the trained regime b) PDF comparison for extrapolated data-sets (prediction) outside the training regime. The model is trained for first $0.1$ sec }
    \label{fig:SDE_drift}
\end{figure}

Fig. \ref{fig:SDE2_drift_a} shows the PDF of the QoI obtained using different methods at $t = 0.03$ sec. We observe that the PDF obtained using the proposed DPC matches exactly with the ground truth. A \textit{data-only}  approach, on the other hand, fails to capture the temporal evolution of the QoI as indicated by the mismatch in the PDF. The performance of the \textit{physics-only} model is better and it correctly captures the mean response. However, it overestimates the  standard deviation obtained as indicated by the flat PDF. This can be attributed to the fact that the drift is considered to be constant in Eq. \eqref{eq:ou_drift} and hence, the diffusion term starts dominating and results in a higher standard deviation.

Fig. \ref{fig:SDE2_drift_b} shows the response PDFs at $t=0.2$ sec. Since the model was trained only up to t=$0.1$ secs, this prediction is outside the  training window. We observe that the proposed DPC yields excellent results with the PDF of QoI matching almost exactly with the ground truth. This clearly indicates the generalization capability of the proposed approach. The \textit{physics-only} and \textit{data-only} models, on the other hand, fail to capture the PDF of the QoI. The failure of the data-driven approach is attributed to its lack of generalization. The \textit{physics-only} model fails because the effect of mis-specified physics becomes prominent with an increase in time.

Next, we consider the case with mis-specified physics in the diffusion. The known physics in this case is represented as
\begin{equation}
      d Y = (x_1 - Y)dt + dB_t
\end{equation}
Other setups, including training data size are kept unchanged.

\begin{figure}[t]
    \centering
    \subfigure[$t=0.08$ sec]{ 
      \includegraphics[width=0.48\textwidth]{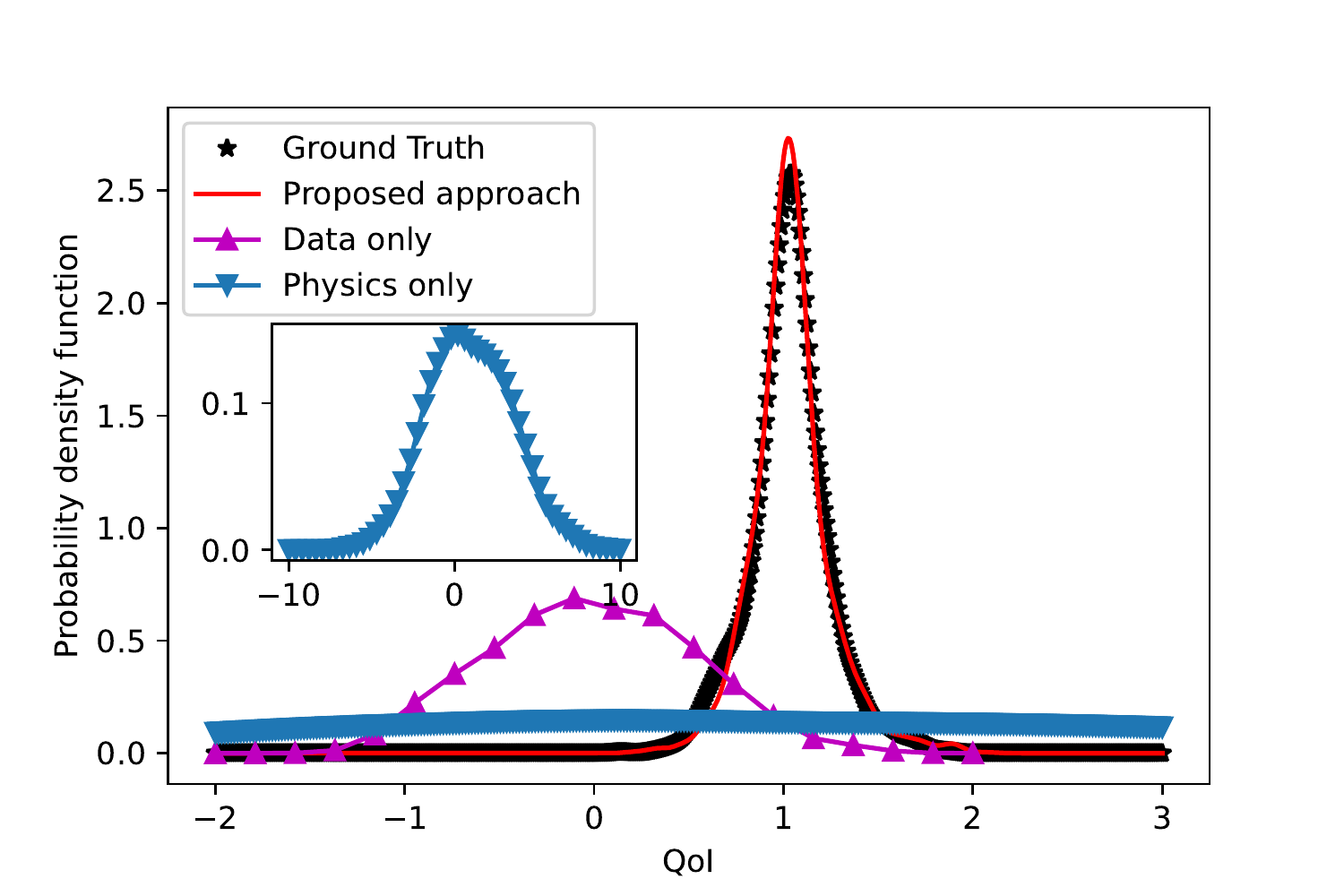}
      \label{fig:SDE2_diff_a}}
    \subfigure[$t=0.59$ sec]{
      \includegraphics[width=0.48\textwidth]{multi_PI_SDE2_PDiff_pred.pdf}
      \label{fig:SDE2_diff_b}}
    \caption{PDF comparison of output data-sets at a specific time, computed from different models with that of the ground truth, for the case where the diffusion term is imprecisely known a) PDF comparison within the trained regime b) PDF comparison for extrapolated data-sets (prediction) outside the training regime. The model is trained for first $0.1$ sec }
    \label{fig:SDE2_diff}
\end{figure}

Figs. \ref{fig:SDE2_diff_a} and  \ref{fig:SDE2_diff_b} show the PDFs of QoI at $t = 0.08$ sec (within the training window) and $t = 0.59$ sec (outside the training window). For both cases, the proposed DPC yields highly accurate results with the response PDF matching exactly with the ground truth. The \textit{physics only} model,
failed to capture the output distribution, with slight offset mean, and completely flat
curve. The offset can be attributed to the accumulated error during time marching, which is
present due to incorrect diffusion term. The \textit{data-only} model also fails to capture the PDF of QoI for both the cases. This indicates poor generalization of purely data-driven approaches.

\subsection{Example 3:  SIR Stochastic model}\label{subsec:eg3}
In this example, we consider Susceptible-Infected-Recovered (SIR) epidemiological model. This model is commonly used to estimate the number of people that are infected with a contagious disease like COVID -19, in a closed population over time. The governing set of SDEs for this model takes the following form:
\begin{equation}
\begin{aligned}
&d S(t) / d t=-\beta S(t) I(t) \\
&d I(t) / d t=\beta S(t) I(t)-\gamma I(t)+\sigma(t, I(t)) d B_{t} \\
&d R(t) / d t=\gamma I(t)
\end{aligned}
\end{equation}
where $S(t)$, $I(t)$ and $R(t)$ denote the number of susceptible cases, infected cases and recovered cases, respectively at any given time $t$. We consider $\alpha = 0.01$, $\beta = 0.5$, $\gamma = 0.5$ and 
\begin{align}
    \sigma(t, I(t)) = \alpha I(t).
\end{align} 
We consider, the total number of infected cases i.e.,  $I(t)$ at any given time t, as the QoI for this example. The initial conditions, $S_0 = S(t=0) \sim \mathcal{U}(1200,1800)$ and $I_0 = T (t=0) \sim \mathcal{U}(20,200)$ are considered to be stochastic. 
Assuming $N(t)$ to be the total population, we have
\begin{align}
    S(t) + I(t) + R(t) = N(t).
\end{align}
We have considered $N(t) = 2000$.


First, we consider the case where the known physics is of the following form:
\begin{equation}\label{eq:SIR_drift}
\begin{aligned}
  &d S(t) / d t=-\beta S(t) I(t) \\
  &d I(t) / d t=-\gamma I(t) + \alpha I(t) d B_{t}\\
  &d R(t) / d t=\gamma I(t)
\end{aligned}
\end{equation}
This corresponds to incomplete/missing physics in the drift component of the SDE. To accommodate for the missing physics, we also have access to high-fidelity sensor data, $\mathcal D = \left\{\bm \xi^{(i)}, \left\{I^{(i)}_{j,1:N_t}\right\}_{j=1}^m \right\}_{i=1}^N$ with $m$ being the number of replication per training sample, $N$ being the number of training samples, and $N_t$ being the time-step until which data is available. Here, $\bm \xi^{(i)} = \left[S_0^{(i)}, I_0^{(i)} \right]$, $m = 50$, $N = 40$, and $N_t = 100$. We have considered a time-step $\Delta t =  0.001$ sec and hence, we have data until $t = 0.1$ sec. The model is trained with a  learning rate of $1\times10^{-6}$, which decreases by a factor of $0.9$ after every 50 epochs. The objective here is to use the proposed DPC to solve the mis-specified physics in Eq. \eqref{eq:SIR_drift} and the high-fidelity data $\mathcal D$.

\begin{figure}[t]
    \centering
    \subfigure[$t=0.01$ sec]{ 
      \includegraphics[width=0.48\textwidth]{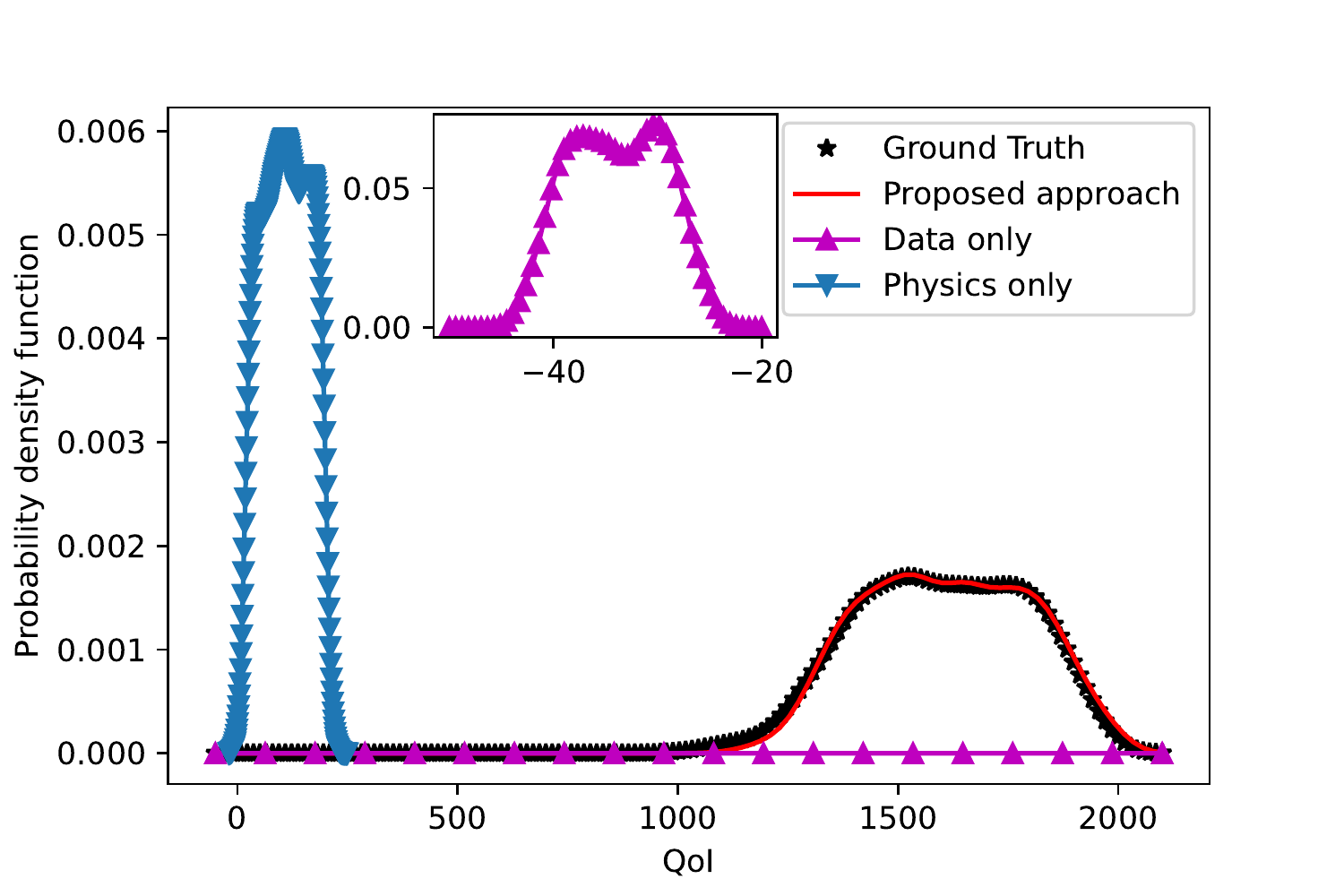}
      \label{fig:SIR_drift_a}}
    \subfigure[$t=0.13$ sec]{
      \includegraphics[width=0.48\textwidth]{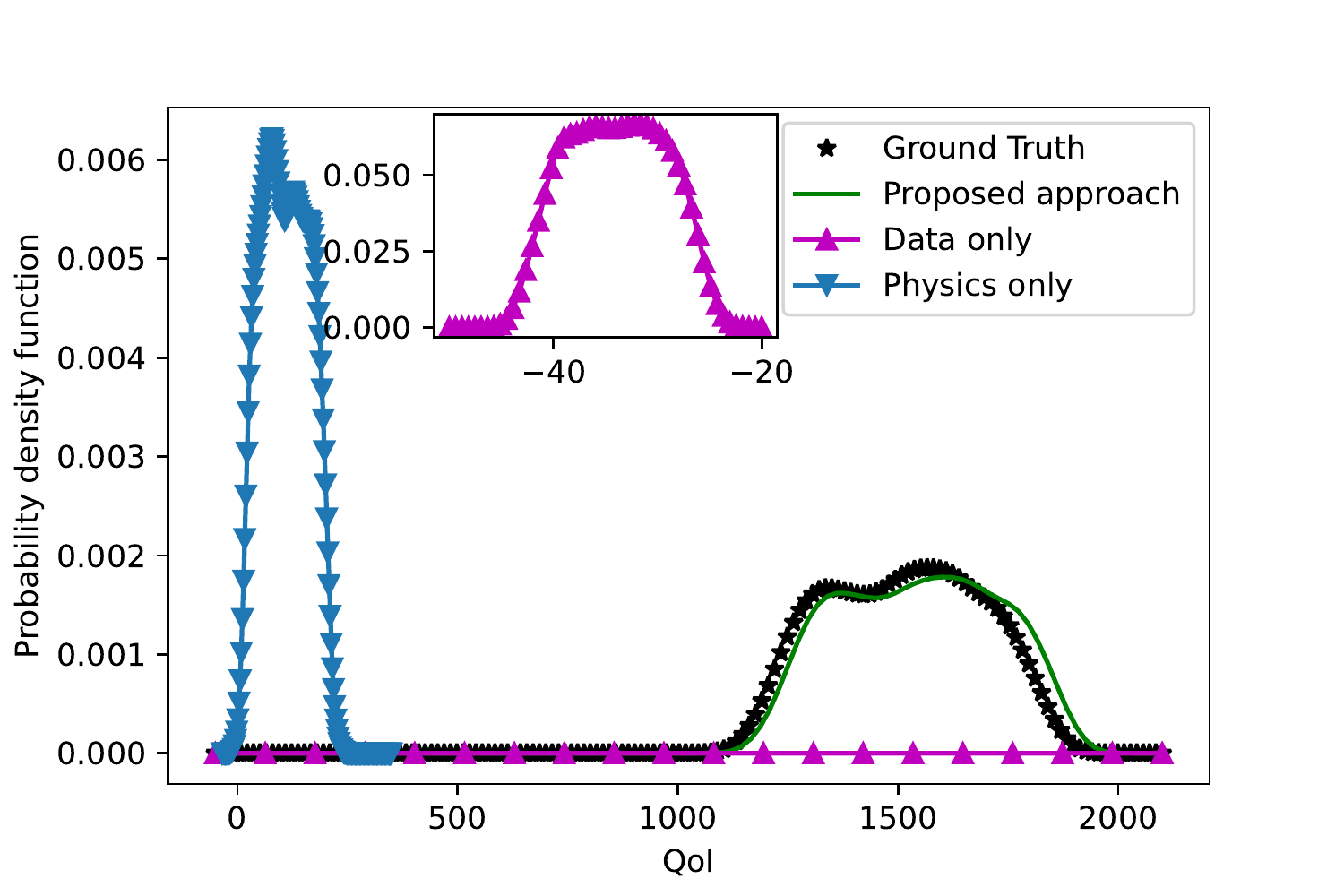}
      \label{fig:SIR_drift_b}}
    \caption{PDF comparison of output data-sets at a specific time, computed from different models with that of the ground truth, for the case where the drift term is imprecisely known a) PDF comparison within the trained regime b) PDF comparison for extrapolated data-sets (prediction) outside the training regime. The model is trained for first $0.1$ sec }
    \label{fig:SIR_drift}
\end{figure}
Fig. \ref{fig:SIR_drift_a} shows the PDF of the QoI obtained using different methods at $t = 0.01$ sec.We observe that the proposed DPC yields very good results, with the PDF of QoI matching almost exactly with that of the ground truth. This indicates the robustness of the proposed approach and its capability in capturing multi-modal distributions.The \textit{physics-only} and the \textit{data-only} models, on the other hand, fail to capture the response PDF.
Fig. \ref{fig:SIR_drift_b} shows the response PDFs at $t=0.13$ sec. Since the model was trained, only up to t=$0.1 secs$, this is a prediction, ahead of the  training window.. The other two models, however, fail to capture the response PDF. The failure of the purely data-driven and purely physics-driven approaches is attributed to the lack of generalization effect of mis-specified physics, respectively.

Next, we consider the case with mis-specified physics in the diffusion. The known physics in this case is represented as
\begin{equation}
  \begin{aligned}
  &d S(t) / d t=-\beta S(t) I(t) \\
  &d I(t) / d t=\beta S(t) I(t) - \gamma I(t) +  d B_{t} \\
  &d R(t) / d t=\gamma I(t)
\end{aligned}
\end{equation}
Other setups, including training data size are kept unchanged.



\begin{figure}
    \centering
    \subfigure[$t=0.03$ sec]{ 
      \includegraphics[width=0.48\textwidth]{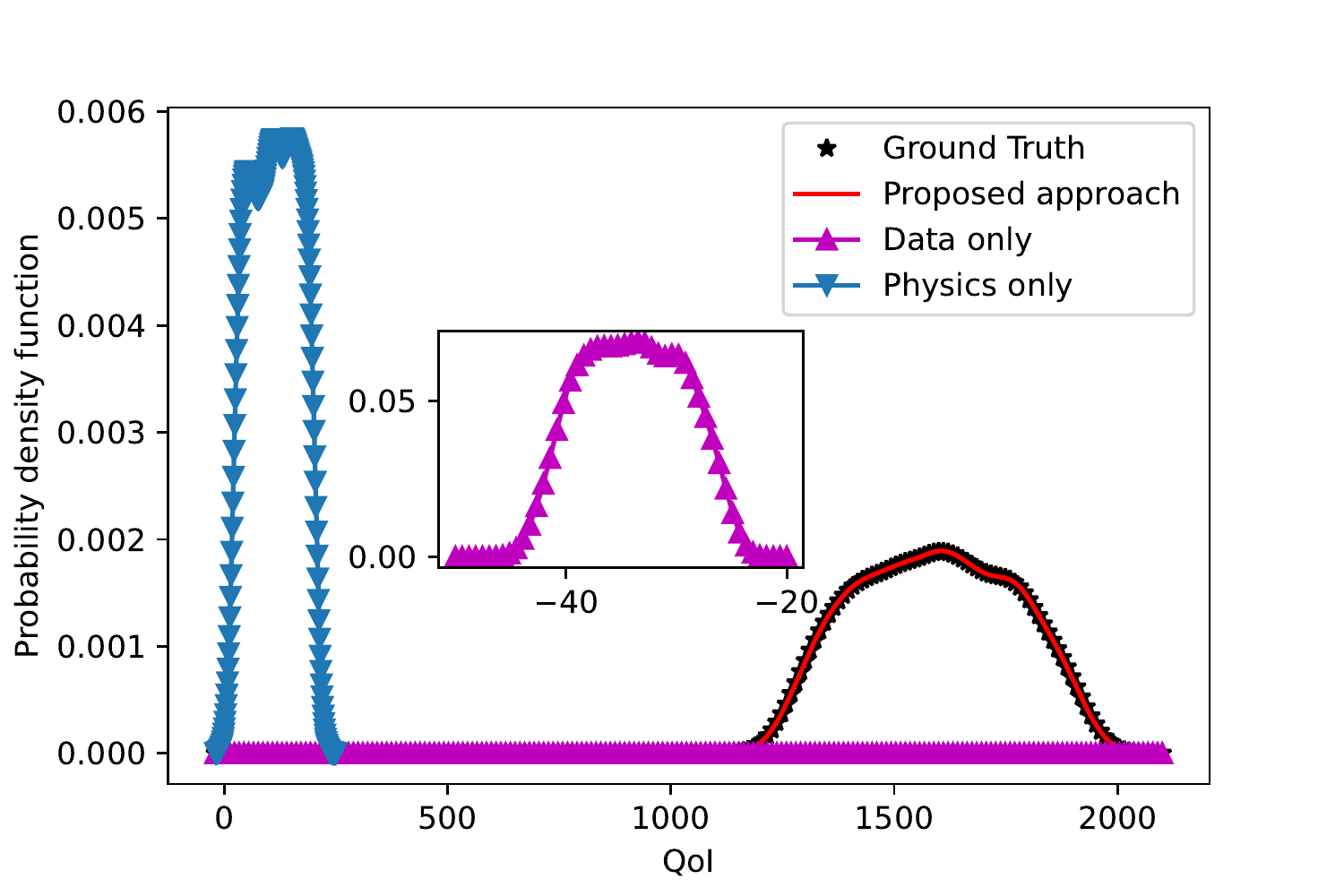}
      \label{fig:SIR_diff_a}}
    \subfigure[$t=0.55$ sec]{
      \includegraphics[width=0.48\textwidth]{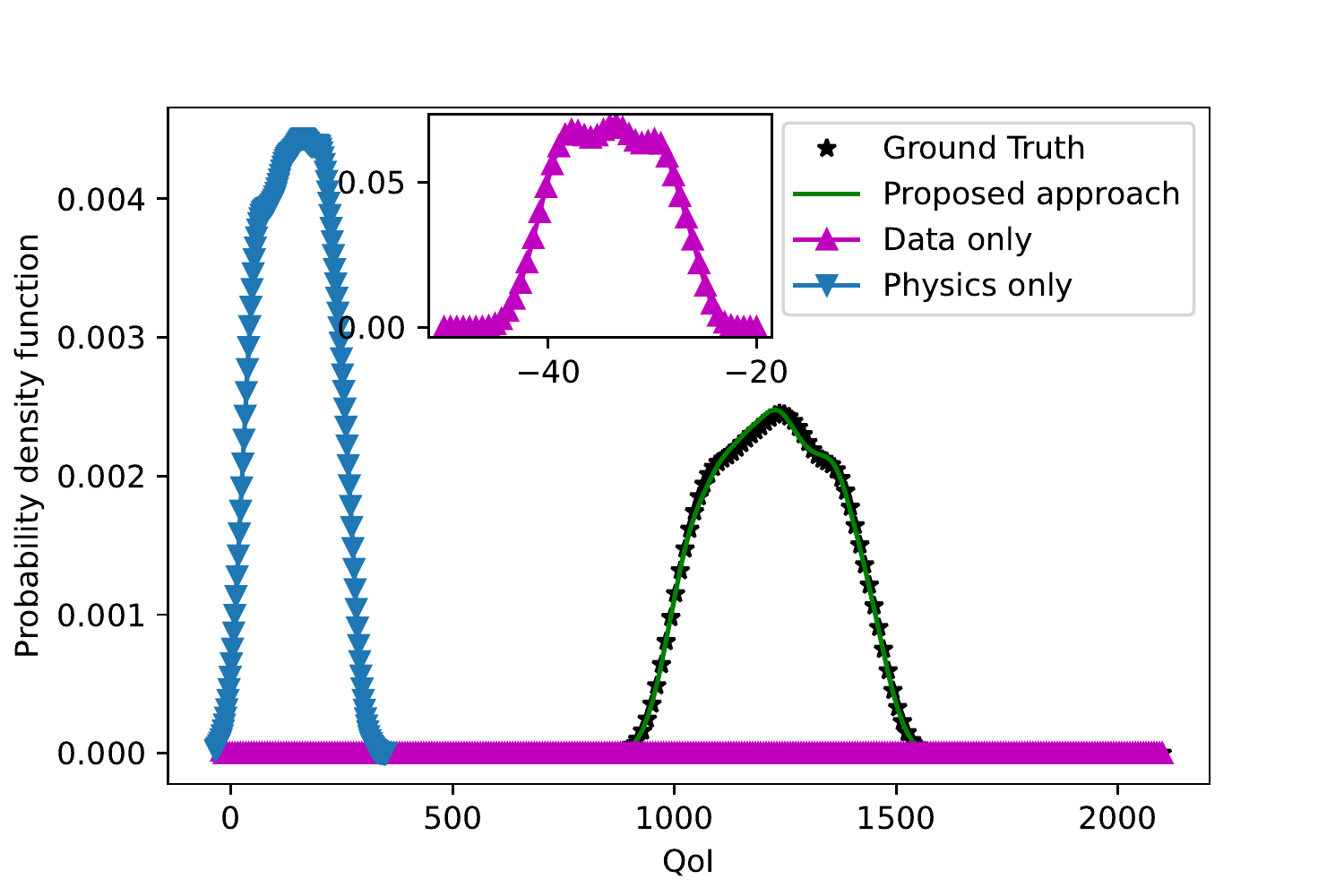}
      \label{fig:SIR_diff_b}}
    \caption{PDF comparison of output data-sets at a specific time, computed from different models with that of the ground truth, for the case where the diffusion term is imprecisely known a) PDF comparison within the trained regime b) PDF comparison for extrapolated data-sets (prediction) outside the training regime. The model is trained for first $0.1$ sec }
    \label{fig:SIR_diff}
\end{figure}

Figs. \ref{fig:SIR_diff_a} and \ref{fig:SIR_diff_b} show the PDFs of QoI at $t=0.03$ sec (within the training window) and  $t=0.55$ sec (outside the training window). For both cases, the proposed DPC yields highly accurate results with the response PDF matching exactly with the ground truth. The \textit{physics-only} model,
failed to capture the output distribution, with a offset in the mean and a sharp peak in the PDF. The offset can be attributed to the accumulated error during time marching, which is
present due to incorrect diffusion term. The \textit{data-only} model also fails to capture the PDF of QoI for both the cases. This indicates poor generalization of purely data-driven approaches.

\subsection{Example 4: Duffing-Van der Pol oscillator}\label{subsec:eg4}
As the last example, we consider stochastic Duffing-Van der Pol oscillator, which is a nonlinear oscillator and has wide applications in the areas of physics, chemistry, biology, engineering and electronics to name a few. We have considered the Duffing-Van der pol oscillator with parametric external excitation \cite{zhou2014chaotic}. The governing SDE for the same looks like:
\begin{equation}\label{eq:DVP}
 \begin{aligned}
  &d X_t = Y_t dt\\
  &dY_t = \left\{\xi_1(1-X_t^2)Y_t + \xi_2 X_t - \alpha X_t^3\right\}dt + \sigma X_t dB_t
 \end{aligned}
 \end{equation}
where $\xi_1 = \frac{c}{k} \sim \mathcal U (0.1, 0.5)$ and $\xi_2 = \frac{k}{m} \sim \mathcal U (5, 50)$are the two input parameters. We consider $\alpha = 100$ and $\sigma = 10^4$. The QoI for this example is taken as the magnitude of displacement $\bm{x}(t)$.

 We first consider the case involving mis-specified drift. We have assumed the term $\xi_1 x^2\cdot\frac{d x}{d t}$ to be zero, thus the known physics is of the following form:
\begin{equation}\label{eq:DVP_drift}
 \begin{aligned}
  &d X_t = Y_t dt\\
  &dY_t = \left\{\xi_1Y_t + \xi_2 X_t - \alpha X_t^3\right\}dt + \sigma X_t dB_t
 \end{aligned}
 \end{equation}
 To accommodate for the missing physics, we also have access to high-fidelity sensor data, $\mathcal D = \left\{\bm \xi^{(i)}, \left\{x^{(i)}_{j,1:N_t}\right\}_{j=1}^m \right\}_{i=1}^N$ with $m$ being the number of replication per training sample, $N$ being the number of training samples, and $N_t$ being the time-step until which data is available. Here, $\bm \xi^{(i)} = \left[\xi_1^{(i)}, \xi_2^{(i)} \right]$, $m = 50$, $N = 40$, and $N_t = 1000$. We have considered a time-step $\Delta t =  0.001$ sec and hence, we have data until $t = 1$ sec. The model is trained with a  learning rate of $1\times10^{-6}$, which decreases by a factor of $0.95$ after every 50 epochs. Similar to the previous examples, we have used ADAM optimizer for training the model. The objective here is to use the proposed DPC to solve the stochastic duffing Van der pol oscillator by utilizing mis-specified physics in  Eq. \eqref{eq:DVP_drift} and the high-fidelity data $\mathcal D$.

\begin{figure}[t]
    \centering
    \subfigure[$t=0.3$ sec]{ 
      \includegraphics[width=0.48\textwidth]{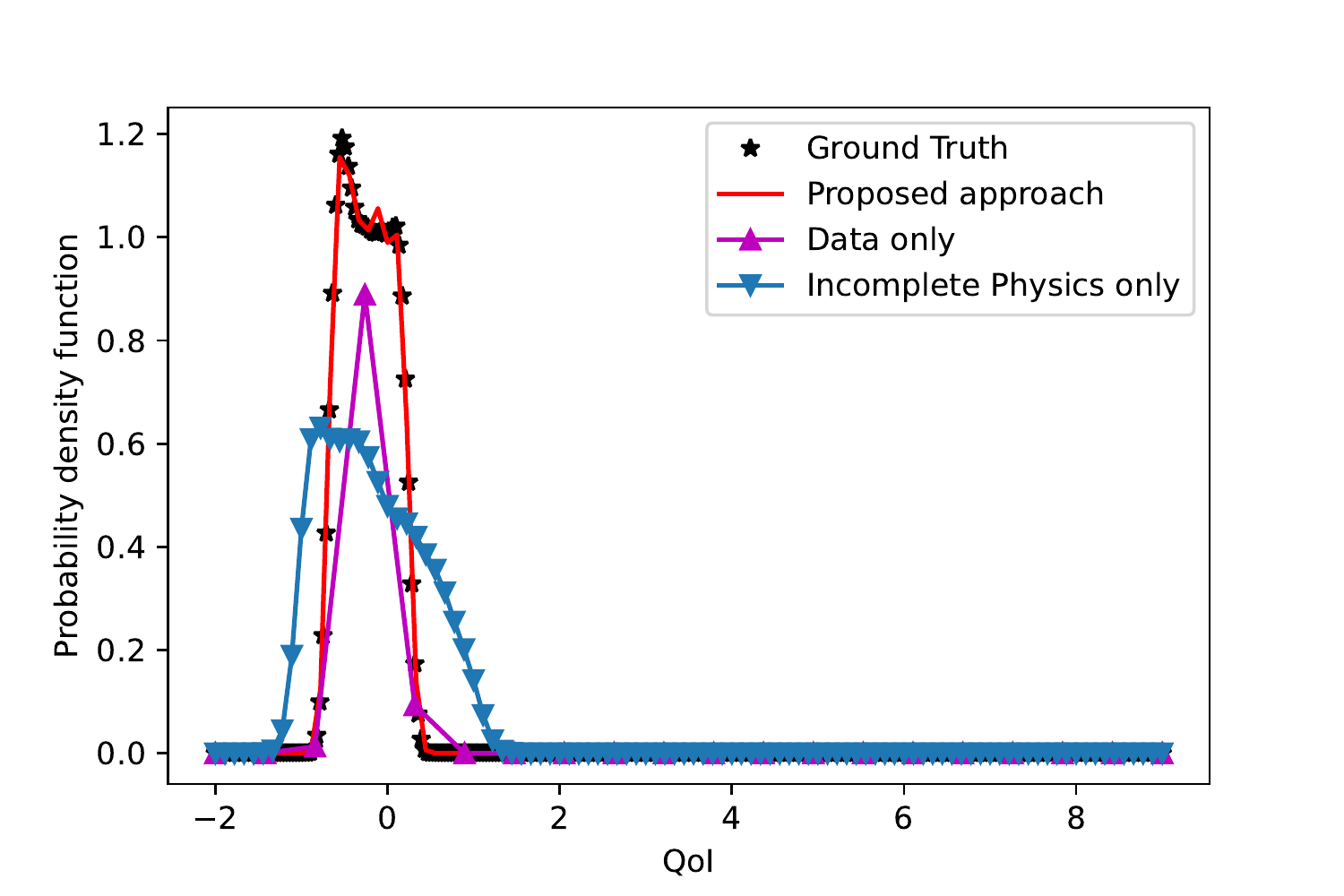}
      \label{fig:DVP_drift_a}}
    \subfigure[$t=8.865$ sec]{
      \includegraphics[width=0.48\textwidth]{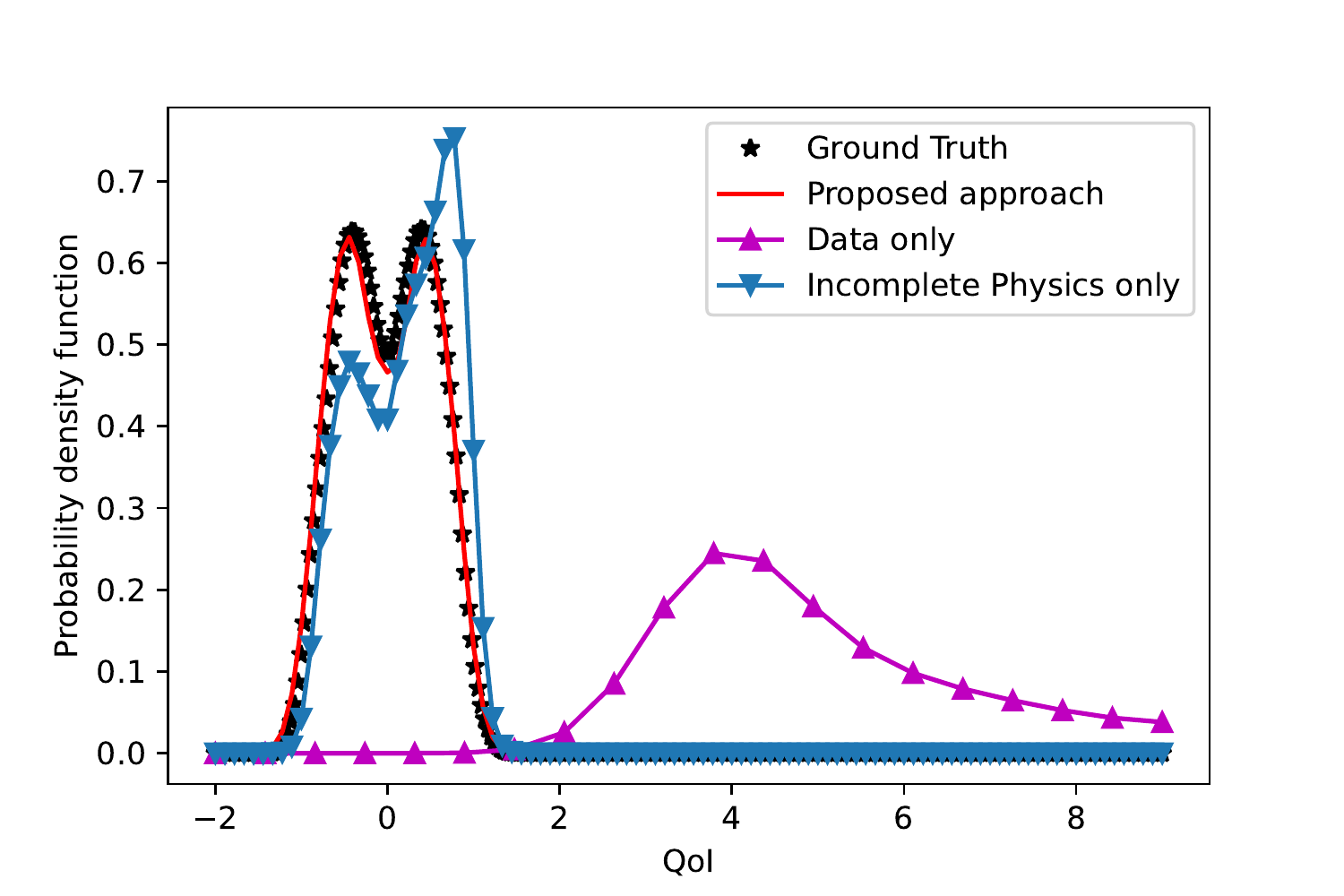}
      \label{fig:DVP_drift_b}}
    \caption{PDF comparison of output data-sets at a specific time, computed from different models with that of the ground truth, for the case where the the drift term is imprecisely known a) PDF comparison within the trained regime b) PDF comparison for predicted data-sets (outside the training window). The model is trained for first $1$ sec }
    \label{fig:DVP_drift}
\end{figure}

 

Fig. \ref{fig:DVP_drift_a} shows the PDF of the QoI obtained using different methods at $t=0.3$ sec. The PDF predicted using the proposed approach matches almost exactly with the ground truth. Methods relying only on physics and only of data, on the other hand, fail to capture the response PDF correctly. In both cases, only a single mode of the response is captured. Fig. \ref{fig:DVP_drift_b} shows the response PDFs at $t=8.865$ secs. Since the model was trained, only up to $t=1$ sec, this is a prediction, way ahead of the  training window. In this case also, the proposed DPC yields excellent results; this illustrates the excellent generalization of the proposed approach. The  \textit{data-only} model fails to capture the PDF of the QoI with offset mean and high standard deviation. This is because the dynamic behavior of duffing Van Der pol is adequately captured by only 1 sec of data.
The physics only model performs satisfactorily in this case. This is because the effect of the missing physics dissipates with time. Nonetheless, results obtained using the proposed approach is best among the three.

Next, we consider the case with mis-specified diffusion. The known physics in this case is represented as
\begin{equation}\label{eq:DVP_diffusion}
 \begin{aligned}
  &d X_t = Y_t dt\\
  &dY_t = \left\{\xi_1Y_t + \xi_2 X_t - \alpha X_t^3\right\}dt + dB_t
 \end{aligned}
\end{equation}
Other setups, including training data size are kept unchanged. We considered a learning rate of $1\times10^{-5}$, which decreases by a factor of 0.95 after every 50 epochs.

\begin{figure}[t]
    \centering
    \subfigure[$t=0.915$ sec]{ 
      \includegraphics[width=0.48\textwidth]{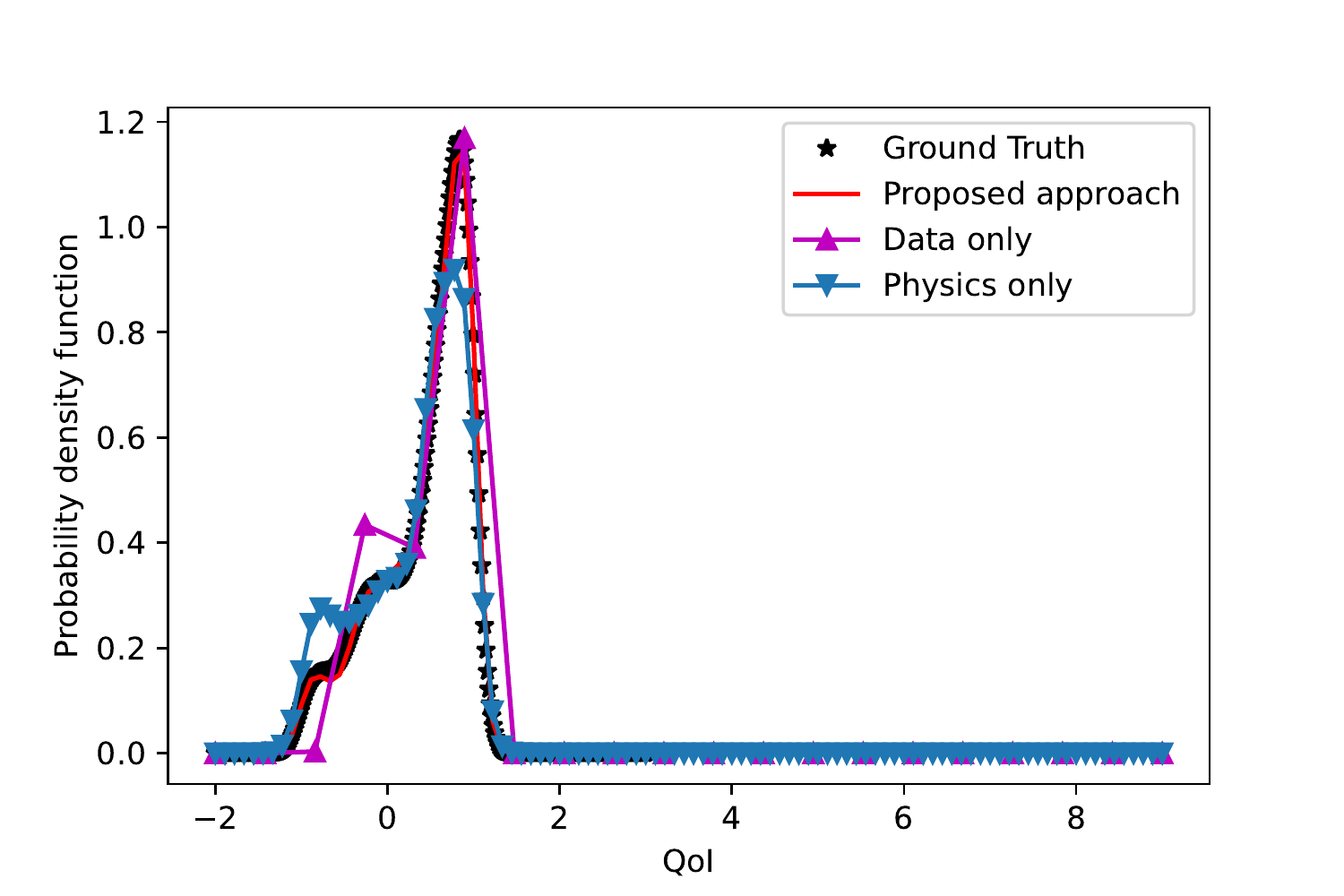}
      \label{fig:DVP_diff_a}}
    \subfigure[$t=8.778$ sec]{
      \includegraphics[width=0.48\textwidth]{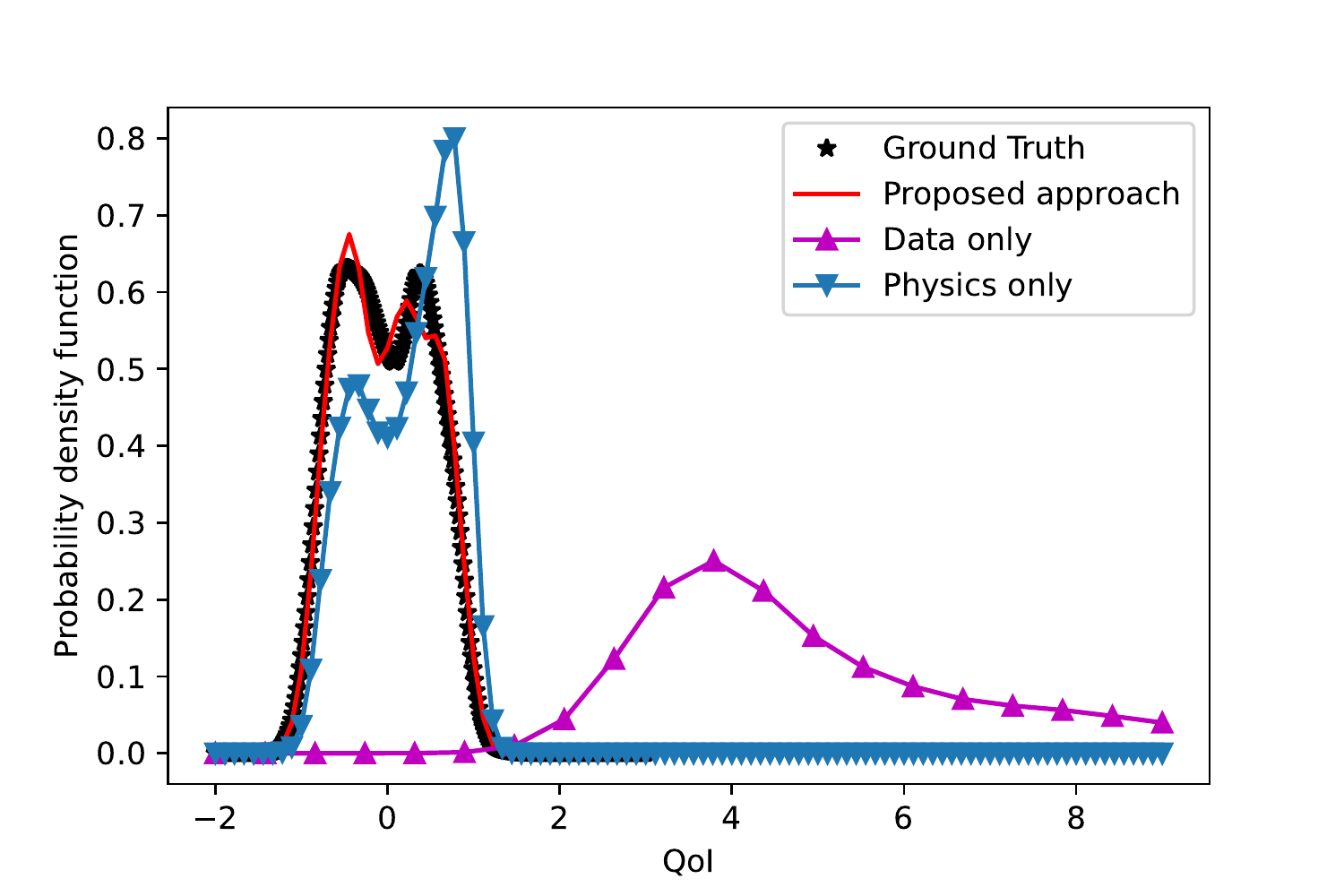}
      \label{fig:DVP_diff_b}}
    \caption{PDF comparison of output data-sets at specific time, computed from different models with that of the ground truth, for the case where the diffusion term is imprecisely known a) PDF comparison within the trained regime b) PDF comparison for predicted data-sets (outside the training window). The model is trained for first $1$ sec }
    \label{fig:DVP_diff}
\end{figure}

Figs. \ref{fig:DVP_diff_a} and \ref{fig:DVP_diff_b} show the PDFs of QoI at $t=0.915$ sec (within the training window) and  $t=8.778$ sec (outside the training window). The proposed DPC yields excellent results in both the cases with response PDFs matching almost exactly with the ground truth. This physics based model, because of the mis-specified diffusion, fails to capture the peaks of the response PDF. Also, the results obtained using physics based model deviates with time. The data only model also captures the response PDF satisfactorily at $t=0.915$ sec. However, it fails of capture the response PDF at $t=8.778$ sec. This again is attributed to the fact that the behavior of the Duffing-Van der Pol oscillator is significantly different outside the training regime.
Overall, based on the examples presented, it is safe to conclude that the proposed approach is robust and can generalize to unseen environments. Finally, before concluding this section, we present a summary of the results elaborated in this section in Table \ref{table:mean_hellinger}.

\begin{table}[ht!]
  \caption{Mean Hellinger distance averaged over time} 
  \label{table:mean_hellinger} 
  \centering 
  \begin{tabular}{c|c|c|c|c|c|c|c|c} 
  \hline\hline
  Method &  \multicolumn{2}{c}{ BS$^{\$\text{\ref{subsec:eg1}}}$} & \multicolumn{2}{|p{2cm}}{MOU$^{\$\text{\ref{subsec:eg2}}}$} & \multicolumn{2}{|p{2cm}}{ SIR$^{\$\text{\ref{subsec:eg3}}}$} & \multicolumn{2}{|p{2cm}}{\centering DVdp$^{\$\text{\ref{subsec:eg4}}}$} \\ \hline
   & Drift & Diffusion  & Drift & Diffusion & Drift & Diffusion & Drift & Diffusion \\
  \hline
  DPC & 0.452 & 0.981 & 0.531 & 0.328 & 0.00482 & 9.86e-6 & 0.0428 & 0.0271\\ 
  Data only & 98.618 & 98.468 & 75.423 & 75.423 &0.227 & 0.299 & 23.811 & 23.764 \\
  Physics only & 66.428 & 61.763 & 58.141 & 45.564 & 0.277 & 0.296 & 1.277 & 1.296\\ [1ex] 
  \hline 
  \end{tabular}
\end{table}

\section{Conclusion}\label{sec:concl}
In this work, we have proposed novel approach, referred to as the Deep Physics Corrector (DPC) for solving stochastic differential equations (SDE). The proposed approach allows seamless integration of known physics and data and hence, is robust and generalizes well to unseen environments. We solved four numerical examples using the proposed approach. The primary observations and key features of the proposed approach are highlighted below:
\begin{itemize}
    \item DPC yields highly accurate results on all the four benchmark example problems, that are carefully selected from different domains, presenting different levels of complexities. This demonstrates the robustness, flexibility and wide applicability of the proposed approach. 
    \item DPC combines approximate physics in such a way so as to retain the advantages of both physics and data. This is evident from the results where the proposed approach yields excellent results even when the test data has significantly different characteristics as compared to the training regime (e.g., in case of Van der pol oscillator).
    \item The proposed approach can be used even when the known physics is significantly different as compared to the actual governing physics. For example, the proposed DPC correctly captured the response PDF for the Black Scholes problem with mis-specified drift and diffusion.
    \item DPC is semi-interpretable in the sense that it incorporates interpretable physics within the network architecture. This is a unique feature of the proposed approach.
    \item The proposed DPC is data efficient. This is evident from the fact that the proposed DPC needed only 0.1 sec of data at a sampling frequency of 1000 Hz for the first three examples and 1 sec of data at sampling frequency of 1000 Hz for the last example. 
\end{itemize}

The Proposed approach has successfully demonstrated its capabilities to work as surrogate stochastic simulator, not only with in the trained region but has also predicted accurate response distribution, way outside it. Having said that the proposed approach has its own limitations. For example, in the current implementation, the number of training samples and training window are selected based on \textit{ad hoc} trial and error approach. A systematic procedure for selecting the training samples is necessary to realize the complete potential of the proposed approach. Also, the proposed framework uses Euler Maruyama scheme for time-integration. Using higher-order stochastic integration scheme can further enhance the framework. In future studies, some of these limitations will be addressed.

\section*{Acknowledgment}
The first author acknowledges the financial support received from Ministry of Education (MOE), IIT Delhi. The second author acknowledges the financial support received from SERB (grant no.: SRG/2021/000467), new faculty seed grant from IIT Delhi, FIRP (grant no.: MI02372) and MFIRP (grant no.: MI02556).

 \end{document}